\documentclass[10pt,twocolumn,letterpaper]{article}

\usepackage{iccv}
\makeatletter
\@namedef{ver@everyshi.sty}{}
\makeatother
\usepackage{tikz}

\usepackage{url}            % simple URL typesetting
\usepackage{booktabs}       % professional-quality tables
\usepackage{amsfonts}       % blackboard math symbols
\usepackage{nicefrac}       % compact symbols for 1/2, etc.
\usepackage{microtype}      % microtypography
\usepackage{xcolor}         % colors

\usepackage{amsmath}
\usepackage{amssymb}
\usepackage{bm}
\usepackage{adjustbox}
\usepackage{dashrule}
\usepackage{mathtools}
\usepackage{makecell}
\usepackage{colortbl}
\usepackage{times}
\usepackage{graphicx}
\usepackage{multirow}

\usepackage{cite} % to order citations
\usepackage{float}
\usepackage{enumitem}
\usepackage{dsfont}
\usepackage{array}
\usepackage{ctable}
\usepackage{comment}
\usepackage{color}
\usepackage{wrapfig, lipsum}
\usepackage{babel}
\usepackage{capt-of}
\usepackage{bigstrut}
\usepackage{arydshln}

\usepackage[accsupp]{axessibility}  % Improves PDF readability for those with disabilities.

\usepackage[
  separate-uncertainty = true,
  multi-part-units = repeat
]{siunitx}

\definecolor{blueblack}{RGB}{0, 108, 173}
\definecolor{taborange}{RGB}{235, 127, 14}
\definecolor{tabgreen}{RGB}{30, 160, 30}
\definecolor{tabpurple}{RGB}{128, 103, 189}
\definecolor{tabred}{RGB}{214, 39, 40}

\usepackage[pagebackref=true,breaklinks=true,bookmarks=false,colorlinks,citecolor=blueblack]{hyperref}

\DeclareMathOperator*{\argmin}{arg\,min}

\definecolor{sol_light_blue}{RGB}{38, 139, 210}
\definecolor{sol_blue}{RGB}{38, 139, 210}
\definecolor{nord_blue}{RGB}{38, 139, 210}
\definecolor{sol_green}{RGB}{163, 190, 140}
\definecolor{sol_red}{RGB}{220, 50, 47}
\definecolor{nord_red}{RGB}{250, 190, 192}
\definecolor{nord_green}{RGB}{163, 190, 140}

\definecolor{beer_orange}{RGB}{242, 142, 28}

\definecolor{nordblack}{RGB}{46, 52, 64}
\definecolor{nordred}{RGB}{191, 97, 106}
\definecolor{magenta}{RGB}{215, 10, 185}
\definecolor{nordgreen}{RGB}{163, 190, 140}
\definecolor{nordblue}{RGB}{94, 129, 172}
\definecolor{nordpurple}{RGB}{180, 142, 160}

\definecolor{realtabgreen}{RGB}{44, 160, 44}
\definecolor{realtabpurple}{RGB}{148, 103, 189}
\definecolor{realtaborange}{RGB}{255, 127, 14}
\definecolor{realtabpink}{RGB}{227, 119, 194}

\def\boxitred#1{%
  \smash{\color{tabred}\fboxrule=1pt\relax\fboxsep=2pt\relax%
  \llap{\rlap{\fbox{\vphantom{0}\makebox[#1]{}}}~}}\ignorespaces
}

\def\boxitgreen#1{%
  \smash{\color{tabgreen}\fboxrule=1pt\relax\fboxsep=2pt\relax%
  \llap{\rlap{\fbox{\vphantom{0}\makebox[#1]{}}}~}}\ignorespaces
}

\newcommand{\bigO}{\mathcal{O}}

\usepackage[capitalize]{cleveref}
\crefname{section}{Sec.}{Secs.}
\Crefname{section}{Section}{Sections}
\Crefname{table}{Table}{Tables}
\crefname{table}{Tab.}{Tabs.}

\iccvfinalcopy % *** Uncomment this line for the final submission

\def\iccvPaperID{6367} % *** Enter the ICCV Paper ID here

\ificcvfinal\fi

\begin{document}

%%%%%%%%% TITLE
\title{Fast Neural Scene Flow}

\author{Xueqian Li$^{*1}$ \:\: Jianqiao Zheng$^{1}$ \:\: Francesco Ferroni$^{*2}$ \:\: Jhony Kaesemodel Pontes$^{*3}$ \:\: Simon Lucey$^{*1}$ \\
\begin{tabular}[h]{cc}
	$^{1}$The University of Adelaide \quad\quad $^{2}$NVIDIA \quad\quad $^{3}$Latitude AI \\
\end{tabular}
}

\twocolumn[{\maketitle
  \ificcvfinal\thispagestyle{empty}\fi
  \centering
  \vspace{-0.2cm}
  \includegraphics[width=0.97\textwidth]{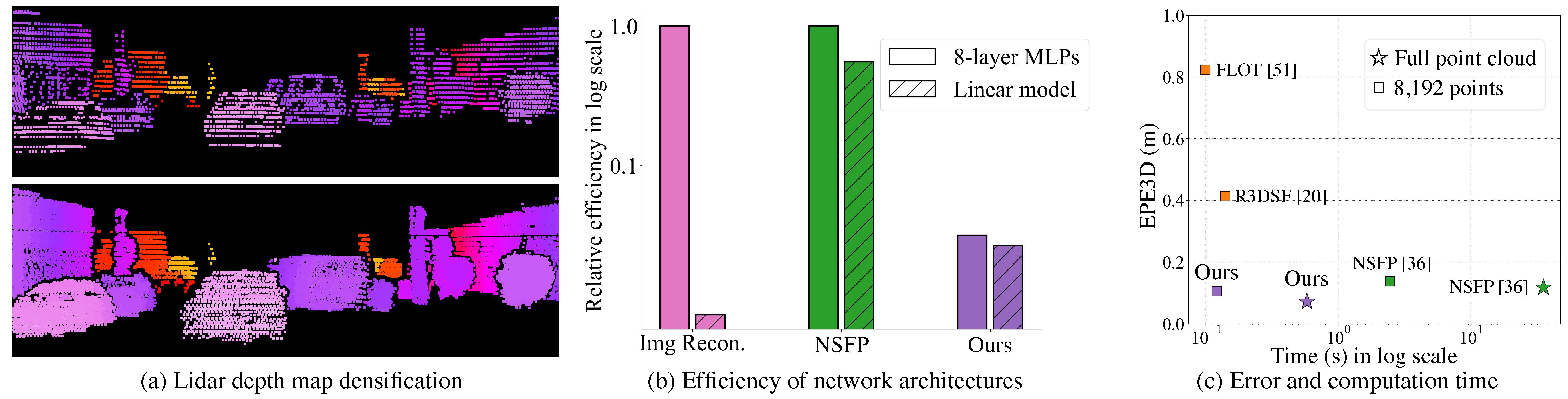}
  \captionof{figure}{
    Scene flow is an important problem as it provides low-level motion cues for many downstream tasks.
    State-of-the-art learning methods are usually fast and can achieve impressive performance on in-domain data, but usually fail to generalize to out-of-the-distribution (OOD) data or handle dense point clouds.
    In this paper, we focus on a runtime optimization-based neural scene flow pipeline. In (a) one can see its application in the densification of lidar. 
    However, in (c) one sees that the major drawback is the extensive computation time.
    We identify that the common speedup strategy in network architectures for coordinate networks has little effect on scene flow acceleration [see \textbf{\color{realtabgreen}green} (b)] unlike image reconstruction [see \textbf{\color{realtabpink}pink} (b)].
    With the dominant computational burden stemming instead from the Chamfer loss function, we propose to use a distance transform-based loss function to accelerate [see \textbf{\color{realtabpurple}purple} (b)], which achieves up to 30$\times$ speedup and on-par estimation performance compared to NSFP [see (c)].
    When tested on 8k points, it is as efficient [see (c)] as leading learning methods, achieving real-time performance.
    }
  \vspace{0.6cm}
\label{fig:figure_01}}]

{
  \renewcommand{\thefootnote}%
    {\fnsymbol{footnote}}
  \footnotetext[1]{Part of the work was done while at Argo AI. 
  Corresponding e-mail: xueqian.li@adelaide.edu.au. Code available at \href{https://github.com/Lilac-Lee/FastNSF.git}{https://github.com/Lilac-Lee/FastNSF.git}}
}

%%%%%%%%% ABSTRACT
\begin{abstract}
    Neural Scene Flow Prior (NSFP) is of significant interest to the vision community due to its inherent robustness to out-of-distribution (OOD) effects and its ability to deal with dense lidar points. 
    The approach utilizes a coordinate neural network to estimate scene flow at runtime, without any training. 
    However, it is up to 100 times slower than current state-of-the-art learning methods. 
    In other applications such as image, video, and radiance function reconstruction innovations in speeding up the runtime performance of coordinate networks have centered upon architectural changes. 
    In this paper, we demonstrate that scene flow is different---with the dominant computational bottleneck stemming from the loss function itself (\ie, Chamfer distance). 
    Further, we rediscover the distance transform (DT) as an efficient, correspondence-free loss function that dramatically speeds up the runtime optimization. 
    Our fast neural scene flow (FNSF) approach reports for the first time real-time performance comparable to learning methods, without any training or OOD bias on two of the largest open autonomous driving (AV) lidar datasets Waymo Open~\cite{sun2020scalability} and Argoverse~\cite{chang2019argoverse}.
\end{abstract}
\vspace{-0.3cm}

% Neural Scene Flow Prior (NSFP) is of significant interest to the vision community due to its inherent robustness to out-of-distribution (OOD) effects and its ability to deal with dense lidar points. The approach utilizes a coordinate neural network to estimate scene flow at runtime, without any training. However, it is up to 100 times slower than current state-of-the-art learning methods. In other applications such as image, video, and radiance function reconstruction innovations in speeding up the runtime performance of coordinate networks have centered upon architectural changes. In this paper, we demonstrate that scene flow is different---with the dominant computational bottleneck stemming from the loss function itself (i.e., Chamfer distance). Further, we rediscover the distance transform (DT) as an efficient, correspondence-free loss function that dramatically speeds up the runtime optimization. Our fast neural scene flow (FNSF) approach reports for the first time real-time performance comparable to learning methods, without any training or OOD bias on two of the largest open autonomous driving (AV) lidar datasets Waymo Open [64] and Argoverse [9].

\section{Introduction}
\label{sc:intro}
Neural Scene Flow Prior (NSFP)~\cite{li2021neural} is considered the dominant method in open-world perception~\cite{najibi2022motion} and scene densification~\cite{li2021neural, wang2022neural} using lidar (see~\cref{fig:figure_01} (a)). 
NSFP achieves state-of-the-art scene flow estimates from dense lidar point clouds (up to 150k+ points) and works on a variety of sensor setups with little to no refinement or adaptation. 
Unlike supervised or unsupervised learning-based methods, NSFP does not require learning from large offline datasets and has no limits on point density ($<$8k for most learning methods). 
Instead, it leverages the architecture of a neural network to implicitly regularize the flow estimate and employs a runtime optimization that can easily scale to large out-of-distribution (OOD) scenes, which is a challenge for current learning-based methods~\cite{pontes2020scene, li2021neural, najibi2022motion, dong2022exploiting, jin2022deformation}.

A fundamental drawback, however, to NSFP is the speed of its runtime optimization which is in some instances of orders of magnitude slower than its learning counterparts (see~\cref{fig:figure_01}). 
As a result, NSFP has widely been used offline for (i) providing scene flow supervision for efficient learning methods and (ii) as a pre-processing step for training open-world perception systems~\cite{najibi2022motion}. 
However, the considerable computational cost of NSFP limits its current applications only to these offline tasks.

A central narrative of our approach is that the dominant computational burden in runtime scene flow optimization (NSFP) is not the network architecture, but the loss function---specifically Chamfer distance (CD)~\cite{fan2017point}. 
This differs considerably from other applications of coordinate networks throughout vision and learning such as neural radiance fields (NeRF~\cite{mildenhall2020nerf}) and high-fidelity image reconstruction (\cite{sitzmann2020implicit}) which have gained significant speedups through architectural innovations~\cite{zheng2022trading}. 
A visual depiction of this discrepancy can be found in~\cref{fig:figure_01} (b).

Key to our approach is the use of correspondence-free loss function---distance transform (DT)~\cite{rosenfeld1966sequential, breu1995linear, danielsson1980euclidean} as a proxy for the computationally expensive CD loss. 
Even though DT has been extensively studied by the graphics and vision community over a few decades, its application as an efficient loss function in deep geometry has largely been overlooked up until this point. 
We believe that the inherent efficiencies of the DT are especially pertinent for runtime network training such as in NSFP. 
Our approach shares similarities with Plenoxels~\cite{yu2021plenoxels}---a recent approach for efficient radiance field estimation using coordinate networks---as we trade memory consumption for computation time, allowing for significant speedups during runtime, which provides an alternative solution when exploring more efficient loss functions for dense scene flow estimation. 
We differ from Plenoxels, however, in that our memory consumption stems from our proposed loss function, not the neural architecture itself.

In this paper, we present for the first time an approximately real-time (for 8k points) runtime optimization method as computationally efficient as leading learning methods whilst preserving the scalability to dense point clouds and state-of-the-art performance on OOD scenes like NSFP.
We compare the performance and the computation time of our approach on two of the largest open lidar AV datasets available: Waymo Open~\cite{sun2020scalability} and Argoverse~\cite{chang2019argoverse}. 
Our fast neural scene flow achieves up to $\sim$30 times speedup than NSFP~\cite{li2021neural} (our faster implementation) and of comparable speed to leading learning-based methods (see~\cref{fig:figure_01} (c)) with the same number of points (8,192).
It opens up the possibility of employing a fast, robust, and generalizable approach for dense scene flow, which is not prone to OOD effects in real-time vision and robotic applications.

\section{Related work}
\label{sc:related}

\vspace{-0.1cm}
\paragraph{Scene flow estimation.}
Scene flow denotes the motion field in 3D space~\cite{vedula1999three} uplifted from 2D optical flow.
To reconstruct 3D flow, traditional image-based methods~\cite{huguet2007variational, li2008multi, hadfield2011kinecting, hadfield2013scene, basha2013multi, quiroga2014dense, hornacek2014sphereflow} formulate an optimization problem utilizing RGB or depth information, and learning-based RGB/RGB-D methods~\cite{shao2018motion, brickwedde2019mono, yang2020upgrading, rishav2020deeplidarflow, teed2021raft, hur2021self, jiang2019sense} rely on single/multiple image features which encode with a large amount of data supervisions.
On the other hand, to estimate scene flow directly from 3D, traditional point cloud-based methods~\cite{chui2003new, pauly2005example, amberg2007optimal} solve for a non-rigid registration problem, while recent work prefers full-supervised learning~\cite{liu2019flownet3d, liu2019meteornet, gu2019hplflownet, wang2020flownet3d++, puy20flot, kittenplon2020flowstep3d, wang2021festa, wu2020pointpwc} that uses point-based features or self-supervised learning~\cite{gojcic2021weakly, wu2020pointpwc, mittal2020just, tishchenko2020self, baur2021slim} that employs a self-supervised loss. 
Recent non-learning-based methods~\cite{pontes2020scene, li2021neural} draw our attention back to runtime optimization that easily scales to large data. 
Graph prior~\cite{pontes2020scene} explicitly builds a graph on the point cloud and uses a graph Laplacian regularizer. 
While neural scene flow prior~\cite{li2021neural} uses the network as an implicit regularizer to smooth motions.
In this paper, we explore point cloud-based scene flow using runtime optimization.

\vspace{-0.35cm}
\paragraph{Accelerating coordinate networks.}
There exists a line of work~\cite{garbin2021fastnerf, reiser2021kilonerf, yu2021plenoctrees, yu2021plenoxels, muller2022instant, hedman2021baking, chen2022tensorf} that focuses on accelerating coordinate networks by trading slow, memory efficient, deep network architectures for fast, memory hungry, shallow architectures. 
Most of these innovations have been applied to the problem of neural radiance fields most notably Plenoxels~\cite{yu2021plenoxels} and TensorRF~\cite{chen2022tensorf}. 
Recently, this trend was generalized for arbitrary signals through the introduction of complex positional encoding~\cite{zheng2022trading} with shallow linear networks. 
In this paper, we claim that these architectural innovations have little utility in speeding up neural scene flow without first addressing the computational cost of the Chamfer loss it uses.

\vspace{-0.35cm}
\paragraph{Distance transform.}
DT~\cite{rosenfeld1966sequential, breu1995linear, danielsson1980euclidean, maurer2003linear, borgefors1996digital} has played an important role in image processing, especially binary image analysis~\cite{niblack1992generating, tran2005efficient}.
Further applications are also found in medical image segmentation~\cite{wang2018deepigeos, wang2020deep, karimi2019reducing, criminisi2008geos, rousson2002shape}, robotics motion planning~\cite{xu2014motion, ratliff2009chomp}, geometric representation~\cite{chan2005level, park2019deepsdf, chen2019learning}, and accelerated point cloud registration~\cite{fitzgibbon2003robust, yang2013go}. 
Among them, various distance measures have been used, such as city block, chessboard, and Euclidean distance~\cite{danielsson1980euclidean, yamada1984complete}. Naturally, Euclidean distance is preferred in computing point distance but it is also the most difficult metric to compute due to the temporal complexity~\cite{grevera2007distance}.
Many work attempts to speedup Euclidean DT computation including raster-scan-based algorithms~\cite{felzenszwalb2012distance, ragnemalm1993euclidean, maurer2003linear, breu1995linear, leymarie1992fast}, fast marching-based algorithms~\cite{mauch2000fast, verwer1989efficient, eggers1998two},~\etc, and has achieved linear time computation.
In this paper, we investigate the raster-scan-based algorithm for the 3D point cloud.

\section{Approach}
\label{sc:approach}

\subsection{Background}
\label{sc:preliminary}

\vspace{-0.1cm}
\paragraph{Scene flow optimization.}

Suppose we have a moving sensor (\eg, lidar mounted on a car, depth camera tied to a robot,~\etc) collecting point cloud in a dynamic environment. 
At time $t\text{-}1$, a point cloud $\mathcal{S}_1$ (source) of the scene is sampled. Then given the movements of the sensor and objects in certain directions, another point cloud $\mathcal{S}_2$ (target) is sampled at time $t$.
In order to find out all the motions in the environment, we model the translation of each point cloud $\mathbf{p} \in \mathcal{S}_1$ from time $t\text{-}1$ to time $t$ as a flow vector $\mathbf{f}\;{\in}\; \mathbb{R}^{3}$, where $\mathbf{p}' = \mathbf{p}+\mathbf{f}$. The translational vector set of all 3D points in $\mathcal{S}_1$ is defined as the scene flow $\mathcal{F}=\{\mathbf{f}_i\}_{i=1}^{|\mathcal{S}_1|}$.

Therefore, the optimization of the scene flow is to minimize the point distance between the source $\mathcal{S}_1$ and the target $\mathcal{S}_2$.
Usually, a regularization $\mbox{C}$, such as a Laplacian regularizer, is needed due to the highly unconstrained non-rigid flows.
The overall optimization becomes
\begin{align}
    \mathcal{F}^* = \argmin_{\mathcal{F}} \sum_{\mathbf{p} \in \mathcal{S}_1} \mbox{D} \left( \mathbf{p}+\mathbf{f}, \mathcal{S}_2 \right) + \lambda \mbox{C}, \label{eq:optim_02}
\end{align}
where $\mbox{D}$ is a point distance function, $\lambda$ is a coefficient of the regularizer $\mbox{C}$.

\vspace{-0.35cm}
\paragraph{Neural scene flow prior.}
NSFP uses traditional runtime optimization to optimize a neural network. 
Different from learning-based methods, NSFP does not rely on any prior knowledge of large-scale datasets. 
And different from traditional scene flow optimization with an explicit regularizer that is mentioned above, neural scene flow prior optimizes parameters of a network which implicitly imposes a regularization by its structure:
\begin{align}
    \mathbf{\Theta}^* = \argmin_{\mathbf{\Theta}} \sum_{\mathbf{p} \in \mathcal{S}_1} \mbox{D} \left( \mathbf{p} + g \left(\mathbf{p}; \mathbf{\Theta} \right), \mathcal{S}_2 \right). \label{eq:optim_main}
\end{align}
where $\mathbf{\Theta}$ is a parameter set of network $g$ to be optimized. $\mathbf{p}$ is the input source point, and the flow $\mathbf{f} \;{=}\; g \left(\mathbf{p};\: \mathbf{\Theta} \right)$ is the output of the network. 
The optimization converges at $\mathbf{f}^{*} \;{=}\; g \left(\mathbf{p};\: \mathbf{\Theta}^{*} \right)$.
The network $g$ here is chosen to be a commonly used ReLU-MLP.

Since the points in source $\mathcal{S}_1$ and target $\mathcal{S}_2$ are not in correspondence, nor having the same number of points,~\ie, $|\mathcal{S}_1| {\neq} |\mathcal{S}_2|$, we use a distance function that handles these problems as
\begin{align}
    \mbox{D} \left(\mathbf{p},\mathcal{S} \right) = \min_{\mathbf{x} \in \mathcal{S}} \lVert \mathbf{p} - \mathbf{x} \rVert_2^2. \label{eq:chamfer_losss}
\end{align}
Practically, a bidirectional point distance is found, yielding the above equation equivalent to the Chamfer loss~\cite{fan2017point}.

\vspace{-0.35cm}
\paragraph{Chamfer distance.}
In point cloud processing, Chamfer distance (CD)~\cite{fan2017point} is an important loss function and metric for computing the point distance of two point clouds that do not necessarily have points in correspondence.
Chamfer distance loss computes the point distance of both source-to-target and target-to-source directions.
In detail, the CD loss can be written as
\begin{align}
    CD (\mathcal{S}_1, \mathcal{S}_2) & = \sum_{\mathbf{p} \in \mathcal{S}_2} \mbox{D} \left(\mathbf{p}, \mathcal{S}_1\right) + \sum_{\mathbf{q} \in \mathcal{S}_1} \mbox{D} \left(\mathbf{q}, \mathcal{S}_2\right) \notag \\ 
    & = \sum_{\mathbf{p} \in \mathcal{S}_2} \min_{\mathbf{x} \in \mathcal{S}_1} \lVert \mathbf{p} - \mathbf{x} \rVert_2^2
    \hphantom{.} + \sum_{\mathbf{q} \in \mathcal{S}_1} \min_{\mathbf{y} \in \mathcal{S}_2} \lVert \mathbf{q} - \mathbf{y} \rVert_2^2.
    \label{eq:cd_expand}
\end{align}
To compute the point distance, correspondences from source-to-target and target-to-source are searched among the nearest point neighbors.
However, the exhaustive point correspondence search is extremely slow, especially when dealing with dense point clouds 
that contain more than 10k points (\eg, in Waymo Open and Argoverse datasets, the number of points can be up to 150k+).

\vspace{0.2cm}
\subsection{Correspondence-free point distance transform}

DT is widely used in 2D image processing, such as segmentation, boundary detection, pattern matching, skeletonization,~\etc. 
However, general usage in irregular and unordered 3D point cloud tasks is not broadly discussed.
Given that the image grid is regular and ordered, a DT map is easily obtained by computing the minimum distance of each sampled point $\mathbf{x}{\in}\mathcal{S}$ to the vertex $\mathbf{q}$ of a DT map $\mathcal{G}$ as
\begin{align}
    \mbox{DT} \left(\mathbf{q} \right) = \min_{\mathbf{x} \in \mathcal{S}, \mathbf{q} \in \mathcal{G}} \mbox{D} \left(\mathbf{x}, \mathbf{q} \right). \label{eq:dt_map_general}
\end{align}
We extend~\cref{eq:dt_map_general} to fit in the scene flow task such that $\mathbf{D}$ refers to Euclidean distance, $\mathbf{x}$ denotes the target point, and $\mathbf{q}$ is the regularly spaced point in a voxel (3D).

\begin{figure}
    \centering
    \includegraphics[width=\linewidth]{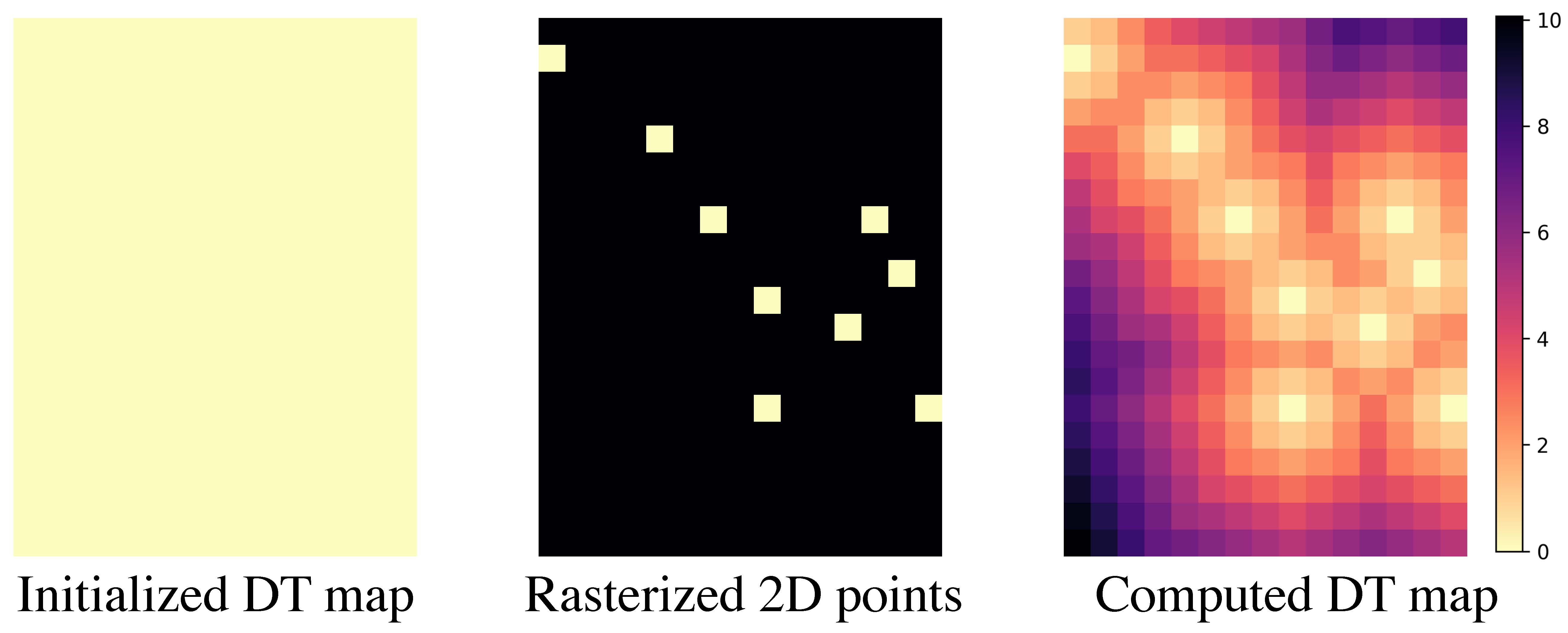}
    \caption{An example of how to build a DT map of 2D points. 
    The initialized DT map is set to all zeros.
    The rasterized 2D point image is set to zero if a point is presented in the grid, and set to one if the grid contains no points.
    After computation, the final DT map is shown in the right figure.
    }
    \label{fig:dt_explain}
    \vspace{-0.3cm}
\end{figure}

\begin{figure}
    \centering
    \includegraphics[width=0.95\linewidth]{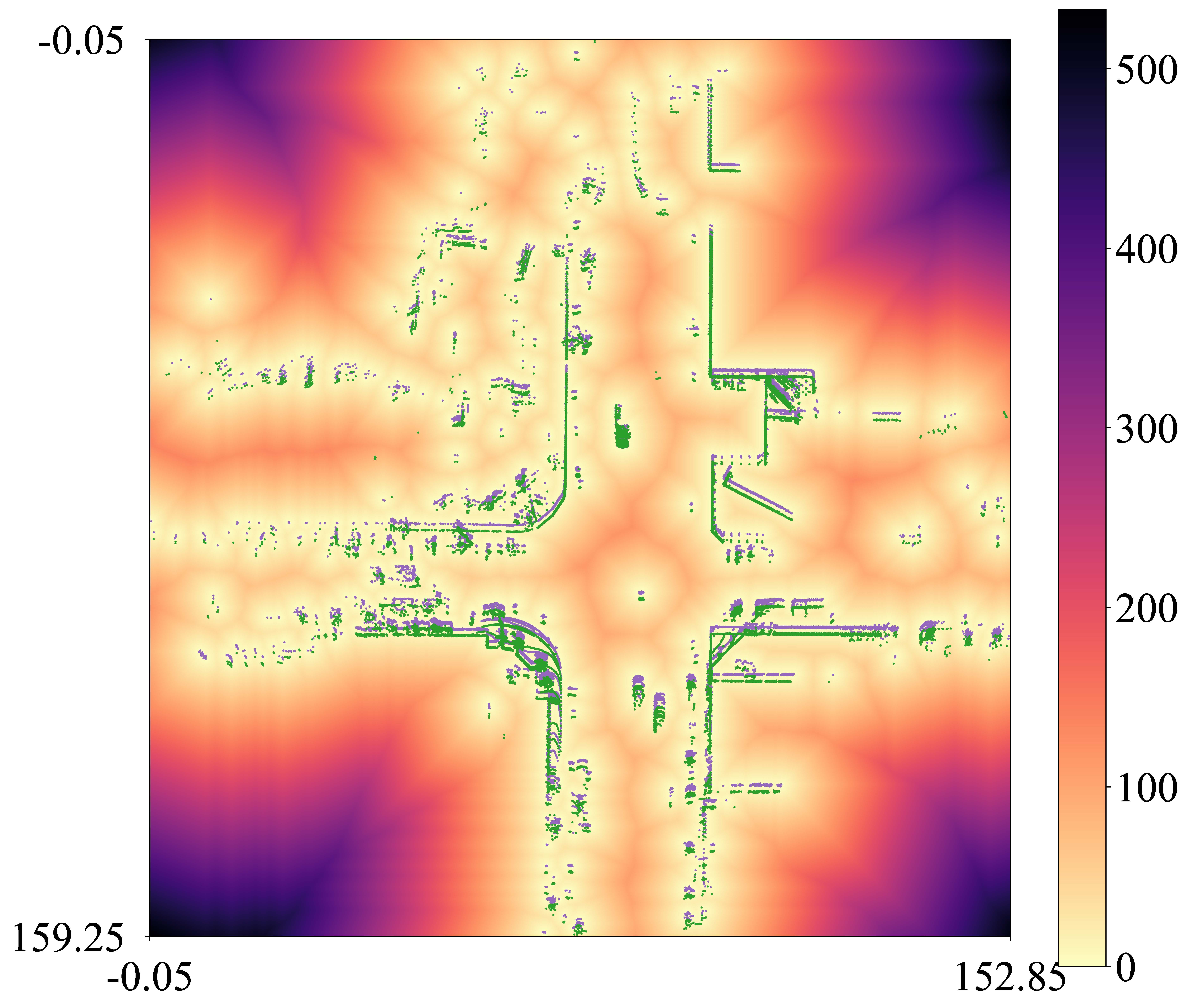}
    \caption{Distance transform map of a target point cloud in Argoverse dataset. 
    Purple points are the target, and green points are the source. 
    It is important to note that the DT map is constructed from the target points, and the source points are only shown for reference purposes.
    The colormap denotes the DT values (in meters), with light yellow denoting a smaller distance and dark purple indicating a larger distance.}
    \label{fig:dt_map}
    \vspace{-0.3cm}
\end{figure}

% \vspace{-0.15cm}
\paragraph{Approximation of DT map.}
However, with a large number of points in a point cloud and the grid map, directly computing~\cref{eq:dt_map_general} builds high dimensional matrices which will lead to large memory occupation.
Even when we presume that each axis of the grid/voxel is separable, and the distance per axis is pre-computed, the memory consumption is still huge that cannot be processed on a single GPU.
One strategy is to iteratively aggregate small matrix multiplications while at a cost of extensive time consumption.
Another strategy is to use k-d tree query to compute the Euclidean distance. Building a k-d tree is one-time computing, but querying from a large point set to an even larger k-d tree is expensive.

Instead, we use a fast distance approximation by rasterizing the irregular points onto the nearest regular cuboid/grid.
Specifically, similar to classical efficient binary image-based DT methods~\cite{felzenszwalb2012distance, criminisi2008geos}, we construct two 3D binary images for target points and an initialized DT map.
An example of 2D DT map is shown in~\cref{fig:dt_explain}.
Specifically, the DT map is set to be all zeros, and the target binary image pixel is set to one when the grid contains no target points in it.
Then a two-pass raster scan-based algorithm~\cite{felzenszwalb2012distance} is applied to each axis to compute the final DT, which means a 3D image needs six propagation passes (left to right, right to left, front to back, back to front, top to bottom, bottom to top).
Although we rasterize target points using a binary 3D image approximation, we still compute the exact Euclidean distance between these two binary images instead of an approximated point distance.
We empirically find that such discretization does not hurt performance.
One reason is the local rigidity of the scene flow benefits from reasonably small but not necessarily infinitesimal DT grids/voxels.

A visualization of a 2D bird's eye view (BEV) DT map is presented in~\cref{fig:dt_map}, where we see that when the point density is large, the distance value is relatively small, and the distance value becomes extremely large when no points exist.
By choosing an appropriate grid size, the pattern of DT is distinct among different points while maintaining a relatively smooth structure in the neighboring area.
Evidently, the correspondence-free distance transform can act as an effective and efficient surrogate for point correspondence-based Chamfer loss.

\vspace{-0.35cm}
\paragraph{DT query and loss function.}
Once a DT map is built for the target point cloud, we can look up the pre-computed DT map to get the distance of the nearest source point in an extremely fast fashion.
The loss function becomes the queried Euclidean distance between the target and the nearest deformed source points:
\begin{align}
    \mathcal{L} = \min_{\mathbf{y} \in \mathcal{S}^{\prime}, \mathbf{q} \in \mathcal{G}} \lVert \mathbf{y} - \mathbf{q} \rVert_2, \label{eq:dt_map_loss}
\end{align}
where $\mathcal{S}^{\prime}$ is the deformed source point cloud.
Unlike CD loss, no point correspondence search is required in~\cref{eq:dt_map_loss}, making the DT query exceptionally fast.
Further ablation studies can be found in~\cref{sc:exp:ablation}.

\section{Experiments}
\label{sc:exp}

% \vspace{-0.05cm}
\paragraph{Datasets}
We are primarily interested in scene flow methods that are well-suited for large-scale, realistic, lidar-based OOD scenes, which are commonly encountered in autonomous driving (AV) applications.
To this end, we focus on two AV datasets: Waymo Open~\cite{sun2020scalability} and Argoverse~\cite{chang2019argoverse}, which contain numerous challenging dynamic scenes.
Unfortunately, no ground truth annotations were provided for the open-world dataset.
We pre-processed the Argoverse and Waymo Open pseudo ground truth scene flow datasets following~\cite{pontes2020scene, li2021neural} and~\cite{yang2021st3d, jin2022deformation} respectively.

\vspace{-0.4cm}
\paragraph{Metrics}
We follow scene flow metrics used in~\cite{liu2019flownet3d, mittal2020just, wu2020pointpwc, pontes2020scene, li2021neural} to evaluate the performance.
We also include the computation time breakdown in the table.
These metrics are: 

\textbf{1) 3D end-point error $\mathcal{E}(m)$} that measures the mean absolute point distance between estimated and target points; 

\textbf{2) strict accuracy $Acc_5(\%)$}, which is the accuracy of the estimated flow that satisfies the absolute point error $\mathcal{E} < 0.05$m or the relative point error $\mathcal{E}^{\prime} < 5\%$; 

\textbf{3) relaxed accuracy $Acc_{10}(\%)$}, which is the accuracy of the estimated flow that satisfies the absolute point error $\mathcal{E} < 0.1$m or the relative point error $\mathcal{E}^{\prime} < 10\%$; 

\textbf{4) angle error $\theta_\epsilon(rad)$} that measures the mean angle error (in radiance) of the estimated and the pseudo ground truth translational vectors. 

\textbf{5) computation time $t (ms)$} breaks down to four parts. \textbf{Pre-compute} includes data loading and building a DT map. 
\textbf{Corr. / DT query} counts the time needed to search point correspondences or query point distance within a DT map. 
We also include computation time for \textbf{Network} forward and backward propagation. 
Finally, \textbf{Total} time in seconds$\hphantom{.}$/$\hphantom{.}$milliseconds is measured.

\paragraph{Implementation details}
We provide the details of implementation for each algorithm we compared. 
Further information can be found in the supplementary materials.

\textbf{1) NSFP~\cite{li2021neural}.} 
The original implementation is extremely slow. 
Based on the official code released by the authors while keeping the same parameter settings specified in the original paper, we implemented a faster version (\textbf{NSFP}) for a fair comparison. 
Note that to reflect the independence of each pair of point clouds when estimating scene flow, we randomly initialized the network before each optimization.

\textbf{2) Baseline.}
Note that although the backward flow enforces a cycle consistency between the deformed point cloud and the original point cloud, we empirically found that removing cycle consistency does not hurt the overall performance when dealing with dense lidar point clouds, but improves the computational efficiency.
Here we removed the backward flow in the original NSFP and implemented a \textbf{NSFP (baseline)} version as our baseline for a fair comparison---all computation times were compared with this model.
We further modified the baseline model using a linear model with complex positional encodings (\textbf{NSFP (baseline, linear)}) to demonstrate the effect of network architecture changes.
Detailed explanations can be found in~\cref{sc:exp:architecture} and the supplementary material.

\textbf{3) Ours.} 
We implemented a DT-based neural scene flow method with 8-layer ReLU-MLPs (\textbf{Ours}). 
We chose the grid cell size of the DT map to be 0.1 meters.
Ablation studies on the choice of grid cell size can be found in~\cref{sc:exp:ablation}.
We further modified our method using a linear model with complex positional encodings (\textbf{Ours (linear)}) to demonstrate the effect of network architecture changes.

\textbf{4) FlowStep3D~\cite{kittenplon2020flowstep3d}} and \textbf{FLOT~\cite{puy20flot}} are fully supervised methods trained on synthetic FlyingThings3D~\cite{mayer2016large} datasets.
\textbf{PointPWC-Net~\cite{wu2020pointpwc}} can be used as a self-supervised method.
We used the official code released by the authors and directly tested the pretrained model (pretrained on FlyingThings3D) on our datasets.
However, as full/self-learning-based methods, they performed poorly on OOD datasets, which is also observed in~\cite{pontes2020scene, li2021neural, najibi2022motion, dong2022exploiting, jin2022deformation}.
Here we only chose the method with the best performance and the lowest computation time---FLOT---for comparison in the main table. The comparison of other learning-based methods is included in the supplementary material.
\textbf{R3DSF~\cite{gojcic2021weakly}} is a weakly supervised method with no direct supervision of ground truth dynamic flows.
We tested the method using the pretrained model (on KITTI~\cite{menze2015object}) provided by the authors.

All models were implemented using CUDA 11.6-supported PyTorch.
All experiments were run on a computer with a single NVIDIA RTX 3090Ti GPU and a 24 AMD Ryzen 9 5900X 12-Core CPU @ 4.95GHz.

\begin{figure}[t!]
    \centering
    \includegraphics[width=\linewidth]{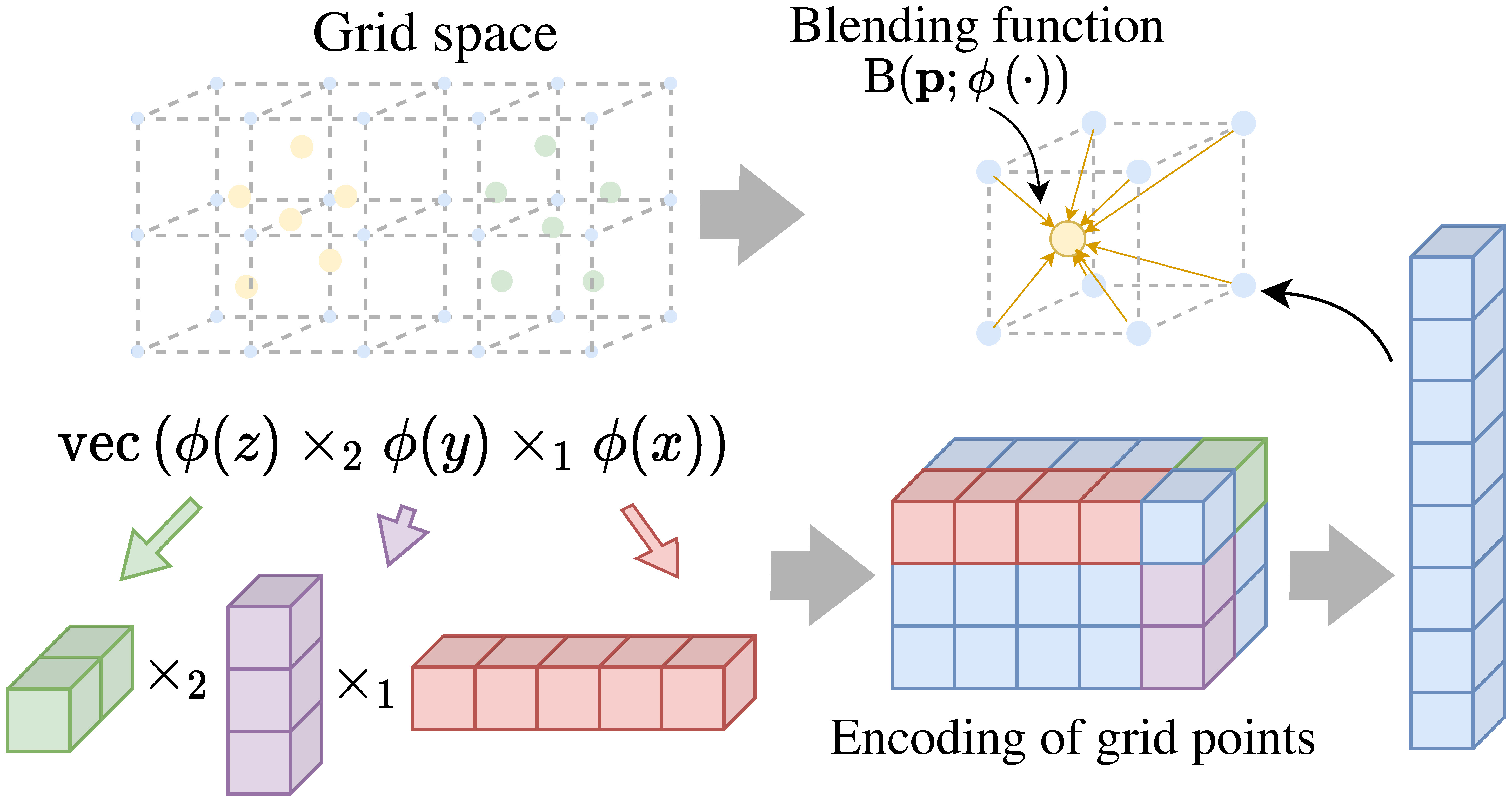}
    \vspace{0.1cm}
    \caption{Illustration of 3D complex PE in scene flow problem. 
    The 3D space of source points (yellow) and target points (green) is divided into small cubes. 
    The number of cubes on each axis differs but the edge length is the same. 
    Encoding of a point in a certain grid is the blending/interpolation of encodings of all 8 grid vertices. 
    These grid points (blue) are encoded by complex PE along each axis. 
    Note the $\mbox{vec}$ and outer product ($\times$) notation is for better visual understanding and it is equivalent to the Kronecker product.}
    \label{fig:3d_complex_encoding}
    \vspace{-0.15cm}
\end{figure}

\begin{table*}[t]
\caption[]{\textbf{Computation time and performance on Waymo Open Scene Flow dataset.} 
We generated 202 testing scene flow examples where each point cloud has 8k-144k points.
The upper tabular between {\color{sol_blue}\textbf{blue bars}} are experiments with the full point cloud, and the lower tabular between {\color{beer_orange}\textbf{orange bars}} are experiments with only 8,192 points.
All query time is listed as time per optimization step [total query time].
The speedups are marked with {\color{tabgreen}green $\times$} numbers, and the slow-downs are marked with {\color{tabred}red $\times$}.
The computation time of using full point cloud is compared to values in the {\color{tabred}red} box, and the computation time of using 8,192 points is compared to values in {\color{tabgreen}green} box.
Bold texts are the best performance, and underlined texts denote the second-best performance.
$\uparrow$ means larger values are better while $\downarrow$ means smaller values are better. 
Corr. / DT query denotes correspondence search or DT query.
\textbf{Note:} performance of learning methods PointPWC-Net~\cite{wu2020pointpwc}, FlowStep3D~\cite{kittenplon2020flowstep3d}, PV-RAFT~\cite{wei2021pv} can be found in the supplementary material.
}
    \centering
    \begin{adjustbox}{width=\linewidth}
    \begin{tabular}{@{}clcccccrccccc@{}}
        \toprule
        &\multirow{2}{*}{\thead{\normalsize Method}}
        &\multirow{2}{*}{\thead{\normalsize ${\mathcal{E}}$\\${(m)}\downarrow$}} 
        &\multirow{2}{*}{\thead{\normalsize ${Acc_5}$\\${(\%)}\uparrow$}} 
        &\multirow{2}{*}{\thead{\normalsize ${Acc_{10}}$\\${(\%)}\uparrow$}} 
        &\multirow{2}{*}{\thead{\normalsize ${\theta_{\epsilon}}$\\${(rad)}\downarrow$}} 
        &\multicolumn{4}{c}{\thead{\normalsize $t~  (ms)\downarrow$}} \\
        \cmidrule{7-10}
        &&&&&& \thead{\normalsize Pre-compute} &\thead{\normalsize Corr. / DT query} & Network &\thead{\normalsize Total} \\
        \midrule
        \arrayrulecolor{sol_blue}\toprule[0.5ex]
        & NSFP~\cite{li2021neural} & 0.100 & \underline{76.62} & \underline{88.56} & \underline{0.286}
        & --- 
        & 114 [30972] $\hphantom{\vert \vert \vert {\color{tabgreen}\mathbf{20{\times}}}}$
        & 4.62 [1361] $\hphantom{\vert \vert \vert .{\color{tabgreen}\mathbf{20{\times}}}}$
        & 35.51 s $\hphantom{\vert 8 8\vert {\color{tabgreen}\mathbf{20{\times}}}}$ \\
        & NSFP (baseline) & 0.118 & 74.16 & 86.70 & 0.300 & --- 
        & \boxitred{3.5in} $\hphantom{.}$ 43.1 [15036] $\hphantom{\vert \vert \vert {\color{tabgreen}\mathbf{20{\times}}}}$
        & 2.38 [904] $\hphantom{\vert \vert \vert .{\color{tabgreen}\mathbf{0{\times}}}}$
        & 18.39 s $\hphantom{\vert 8 8\vert {\color{tabgreen}\mathbf{20{\times}}}}$ \\
        & NSFP (baseline, linear) & \underline{0.096} & 70.78 & 86.31 & 0.310 
        & 9.48 %$\hphantom{\vert {\color{tabred}\mathbf{7e{-}4{\times}}}}$ 
        & 40.38 [8037] $\vert \hphantom{\vert}{\color{tabgreen}\mathbf{1.1{\times}}}$ 
        & 1.53 [319] $\vert {\color{tabgreen}\mathbf{1.6{\times}}}$ 
        & 10.20 s $ \vert {\color{tabgreen}\mathbf{1.80{\times}}} \hphantom{1}$ \\
        & Ours & \textbf{0.072} & \textbf{84.73} & \textbf{92.24} & \textbf{0.280} 
        & 40.85 %$\vert {\color{tabred}\mathbf{0.09}\times}$ 
        & 0.25 [13] $\vert {\color{tabgreen}\mathbf{172{\times}}}$ 
        & 2.89 [149] $\vert {\color{tabred}\mathbf{1.2{\times}}}$ 
        & \underline{0.58 s} $\vert {\color{tabgreen}\mathbf{31.7{\times}}}$ \\
        & Ours (linear) & 0.109 & 71.27 & 85.80 & 0.321
        & 44.88 %$\vert {\color{tabred}\mathbf{7e{-}4{\times}}}$
        & 0.23 [19] $\vert {\color{tabgreen}\mathbf{{187\times}}}$ 
        & 1.62 [138] $\vert {\color{tabgreen}\mathbf{{1.5\times}}}$ 
        & \textbf{0.49 s} $\vert {\color{tabgreen}\mathbf{37.5{\times}}}$ \\
        \arrayrulecolor{sol_blue}\toprule[0.5ex]
        \arrayrulecolor{beer_orange}\toprule[0.5ex]
        & FLOT~\cite{puy20flot} & 0.702 & 2.46 & 11.30 & 0.808 & --- 
        & --- $\hphantom{----}$ & --- 
        & 99 ms $\hphantom{h}$ $\hphantom{\vert {\color{tabred}\mathbf{.1}}}$ $\hphantom{hh}$ \\
        & R3DSF~\cite{gojcic2021weakly} & 0.414 & 35.47 & 44.96 & 0.527 & --- & --- $\hphantom{----}$ & --- 
        & \boxitgreen{1.01in} \underline{140 ms} $\hphantom{h}$ $\hphantom{\vert {\color{tabred}\mathbf{1.4{\times}}}}$ $\hphantom{.}$ \\
        & NSFP~\cite{li2021neural} (8,192 pts) & \underline{0.138} & \underline{53.62} & \underline{78.57} & \underline{0.339} & --- 
        & 5.42 [1285] $\vert \hphantom{\vert\vert} {\color{tabgreen}\mathbf{21{\times}}}$
        & 4.44 [1051] $\vert \hphantom{\vert} {\color{tabgreen}\mathbf{1.1{\times}}}$
        & 2459 ms $\vert {\color{tabred}\mathbf{17.6{\times}}}$ $\hphantom{|..}$ \\
        & Ours (8,192 pts) & \textbf{0.106} & \textbf{77.53} & \textbf{88.99} & \textbf{0.329} & 35.22 
        & 0.23 [6.5] $\vert {\color{tabgreen}\mathbf{496{\times}}}$
        & $\hphantom{||}$ 2.60 [76] $\vert {\color{tabgreen}\mathbf{1.8{\times}}}$
        & \textbf{121 ms} $\vert {\color{tabgreen}\mathbf{1.16{\times}}}$ $\hphantom{.}$ \\
        \arrayrulecolor{beer_orange}\toprule[0.5ex]
        \arrayrulecolor{black}\bottomrule
    \end{tabular}
    \end{adjustbox}
    \label{tab:mean_time_waymo_3d}
    \vspace{-0.1cm}
\end{table*}

\begin{table*}[t]
\caption[]{\textbf{Computation time and performance on Argoverse Scene Flow dataset.} Argoverse has 212 testing scene flow examples where each point cloud has 30k-70k points.
Notations are the same as in the table above.
}
    \centering
    \begin{adjustbox}{width=\linewidth}
    \begin{tabular}{@{}clcccccrccccc@{}}
        \toprule
        &\multirow{2}{*}{\thead{\normalsize Method}}
        &\multirow{2}{*}{\thead{\normalsize ${\mathcal{E}}$\\${(m)}\downarrow$}} 
        &\multirow{2}{*}{\thead{\normalsize ${Acc_5}$\\${(\%)}\uparrow$}} 
        &\multirow{2}{*}{\thead{\normalsize ${Acc_{10}}$\\${(\%)}\uparrow$}} 
        &\multirow{2}{*}{\thead{\normalsize ${\theta_{\epsilon}}$\\${(rad)}\downarrow$}} 
        &\multicolumn{4}{c}{\thead{\normalsize $t~(ms)\downarrow$}} \\
        \cmidrule{7-10}
        &&&&&& \thead{\normalsize Pre-compute} &\thead{\normalsize Corr. / DT query} & Network &\thead{\normalsize Total} \\
        \midrule
        \arrayrulecolor{sol_blue}\toprule[0.5ex]
        & NSFP~\cite{li2021neural} & \textbf{0.069} & \underline{71.56} & \underline{87.80} & \textbf{0.235}
        & --- 
        & 47 [14310] $\hphantom{\vert \vert \vert {\color{tabgreen}\mathbf{20{\times}}}}$
        & 4.66 [1507] $\hphantom{\vert \vert \vert .{\color{tabgreen}\mathbf{20{\times}}}}$
        & 18.08 s $\hphantom{\vert 8 8\vert {\color{tabgreen}\mathbf{20{\times}}}}$ \\
        & NSFP (baseline) & 0.078 & 69.46 & 86.22 & \underline{0.253}
        & --- 
        & \boxitred{3.5in} $\hphantom{....}$ 17 [5901] $\hphantom{\vert \vert \vert {\color{tabgreen}\mathbf{20{\times}}}}$
        & 2.31 [848] $\hphantom{\vert \vert {\color{tabgreen}\mathbf{20{\times}}}}$
        & 8.38 s $\hphantom{\vert 8 \vert {\color{tabgreen}\mathbf{20{\times}}}}$ \\
        & NSFP (baseline, linear) & 0.097 & 67.03 & 83.20 & 0.314 
        & 9.19 %$\hphantom{\vert {\color{tabred}\mathbf{7e{-}4{\times}}}}$ 
        & 14.7 [2786] $\vert \hphantom{.} {\color{tabgreen}\mathbf{1.1{\times}}}$ 
        & 1.51 [297] $\vert {\color{tabgreen}\mathbf{1.5{\times}}}$ 
        & 3.55 s $\vert {\color{tabgreen}\mathbf{2.36{\times}}}$ \\
        & Ours & \underline{0.071}& \textbf{80.05} & \textbf{90.71} & 0.289 
        & 43.61 %$\vert {\color{tabred}\mathbf{0.09}\times}$ 
        & 0.24 [14] $\vert \hphantom{..} {\color{tabgreen}\mathbf{71{\times}}}$ 
        & 2.57 [149] $\vert {\color{tabred}\mathbf{1.1{\times}}}$ 
        & \underline{0.51 s} $\vert {\color{tabgreen}\mathbf{16.4{\times}}}$ \\
        & Ours (linear) & 0.106 & 65.00 & 82.85 & 0.319
        & 48.59 %$\vert {\color{tabred}\mathbf{7e{-}4{\times}}}$
        & 0.23 [20] $\vert \hphantom{..} {\color{tabgreen}\mathbf{{74\times}}}$ 
        & 1.69 [149] $\vert {\color{tabgreen}\mathbf{1.4{\times}}}$ 
        & \textbf{0.43 s} $\vert {\color{tabgreen}\mathbf{19.5{\times}}}$ \\
        \arrayrulecolor{sol_blue}\toprule[0.5ex]
        \arrayrulecolor{beer_orange}\toprule[0.5ex]
        & FLOT~\cite{puy20flot} & 0.821 & 2.00 & 8.84 & 0.967 & --- 
        & --- $\hphantom{----}$ & --- 
        & 88 ms $\hphantom{h}$ $\hphantom{\vert {\color{tabred}\mathbf{{.1}}}}$ $\hphantom{.}$ \\
        & R3DSF~\cite{gojcic2021weakly} & 0.417 & 32.52 & 42.52 & 0.551 & --- & --- $\hphantom{----}$ & --- 
        & \boxitgreen{1.01in} $\hphantom{h}$ \textbf{113 ms} $\hphantom{h}$ $\hphantom{\vert {\color{tabred}\mathbf{1.4{\times}}}}$ $\hphantom{.}$ \\
        & NSFP~\cite{li2021neural} (8,192 pts) & \textbf{0.113} & \underline{46.32} & \underline{72.68} & \textbf{0.347} & --- 
        & 5.40 [1500] $\vert \hphantom{\vert} {\color{tabgreen}\mathbf{8.7{\times}}}$
        & 4.42 [1233] $\vert \hphantom{\vert} {\color{tabgreen}\mathbf{1.1{\times}}}$
        & 2864 ms $\vert {\color{tabred}\mathbf{25.6{\times}}}$ $\hphantom{.}$ \\
        & Ours (8,192 pts) & \underline{0.118} & \textbf{69.93} & \textbf{83.55} & \underline{0.352} & 41.57 
        & 0.22 [6.33] $\vert {\color{tabgreen}\mathbf{214{\times}}}$
        & 2.51 [72.69] $\vert \hphantom{\vert} {\color{tabgreen}\mathbf{1.9{\times}}}\hphantom{.}$
        & $\hphantom{|}$ \underline{124 ms} $\vert {\color{tabred}\mathbf{1.10{\times}}}$ $\hphantom{.}$ \\
        \arrayrulecolor{beer_orange}\toprule[0.5ex]
        \arrayrulecolor{black}\bottomrule
    \end{tabular}
    \end{adjustbox}
    \label{tab:mean_time_argoverse_3d}
   \vspace{-0.2cm}
\end{table*}

\subsection{Speedup in network architecture}\label{sc:exp:architecture}
Plenoxels~\cite{yu2021plenoxels} is a method that accelerates NeRF~\cite{mildenhall2020nerf} optimization by replacing deep neural networks with voxel-based spherical harmonics representations, leading to efficient volume rendering.
Similar network speedup approaches can also be used in neural scene flow optimization.
We incorporate recent innovations in positional encodings, specifically complex positional encoding (complex PE)~\cite{zheng2022trading} to represent high-frequency signals and enable a linear reconstruction model which is similar in spirit to~\cite{yu2021plenoxels}.

\vspace{-0.35cm}
\paragraph{Complex positional encodings}
Positional encodings (PEs) are encodings for input positions---\eg, 2D image grids, 3D voxel grids,~\etc.---that are usually used in coordinate networks.
PEs have been shown to improve the performance and convergence speed of coordinate networks~\cite{tancik2020fourier}.
Simple PE is a simple concatenation of the encoding in each input dimension, while a complex PE is a more complicated encoding that computes the Kronecker product of the per-dimension encoding.
An illustration of complex PE of scene flow is shown in~\cref{fig:3d_complex_encoding}.

With a complex PE and a linear model parameterized by $\mathbf{W} \:{\in}\: \mathbb{R}^{W_xW_yW_z \times 3}$, the scene flow can be represented as
\begin{align}
    \mathbf{f} \approx \mbox{B}(\mathbf{p};\phi) \mbox{vec} \left(\phi(\mathbf{x}) \mathbf{W} \phi(\mathbf{z})^T \phi(\mathbf{y})^T \right), \label{eq:complex_encoding}
\end{align}
where $\mbox{B}(\cdot)$ is the blending function, $\phi\left(\cdot\right)$ is the encoder, $\phi(\mathbf{x}) \mathbf{W} \phi(\mathbf{z})^T \phi(\mathbf{y})^T$ is a simple notation for $n$-mode multiplication, which is equivalent to Kronecker product.
The blending function interpolates the 3D regular grid encodings to handle the irregular and unordered nature of 3D point clouds.
Additional details of complex PE can be found in the supplementary materials.

\paragraph{Network speedup comparison}
We would like to point out that although various strategies have been proposed to accelerate network architectures, they do not lead to substantial speedup in neural scene flow estimation.
To demonstrate this, we compare the relative efficiency of using 8-layer Relu-MLPs and complex PE with a linear model in both 2D image reconstruction and 3D scene flow tasks, shown in~\cref{fig:figure_01} (b).
In detail, following~\cite{zheng2022trading}, we used a size of 256$\times$256 image dataset for image reconstruction.
For a fair comparison, the relative efficiency is obtained by normalizing the computation time of two tasks.
The computation time of single image reconstruction is 17.63 s and 0.15 s for deep network and linear network respectively, leading to 118$\times$ speedup.
However, the linear model only results in $\sim$2$\times$ and $\sim$1.2$\times$ speedup for NSFP and our method respectively. 
Our results show that complex PE-based speedup achieves significant acceleration in 2D image reconstruction compared to deep neural networks, but the benefits of such speedup are more limited when estimating neural scene flow.
More detailed comparisons are in~\cref{sc:exp:benchmark}.

\begin{figure*}[t]
\centering
    \includegraphics[width=0.95\linewidth]{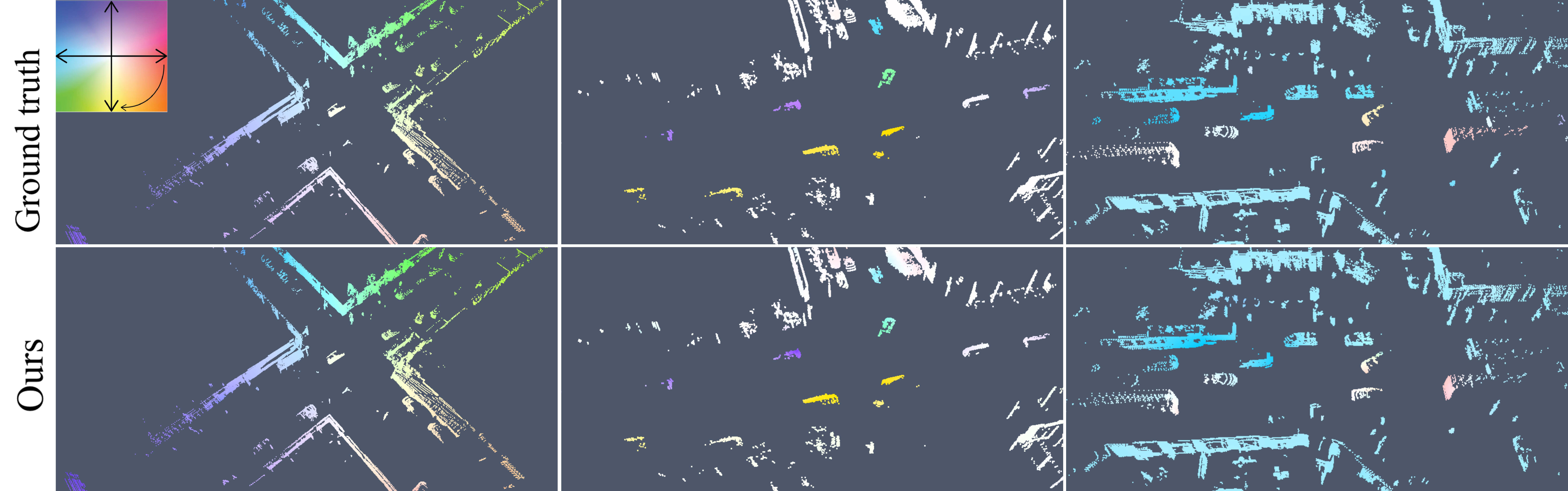}
    \caption{Visual examples of the scene flow prediction using our method on the Argoverse dataset. 
    We show 3 different autonomous driving scenarios where the autonomous vehicle (AV) was making a right turn (left), the AV stopped for crossing traffic (middle), and the AV was driving in the city (right). 
    We found that the scene flow predicted by our method is close to the ground truth flows.
    The upper left corner color wheel indicates the flow magnitude (color intensity) and the flow direction (angle).
    For example, the yellow vehicles in the middle figure are heading south with a relatively large speed \textit{wrt} the AV.} 
    \label{fig:qualitative}
    \vspace{-0.3cm}
\end{figure*}

\subsection{Comparison of performance}\label{sc:exp:benchmark}

We show the performance of our method with different variants compared to NSFP, FLOT, and R3DSF on Waymo Open (\cref{tab:mean_time_waymo_3d}) and Argoverse (\cref{tab:mean_time_argoverse_3d}) scene flow datasets. 
We denote results on Waymo Open (xx) and Argoverse (yy) as xx/yy.
Visual results and applications of densification are shown in~\cref{fig:qualitative} and supplementary materials.

\vspace{-0.32cm}
\paragraph{Dense scene flow estimation.}

For dense point clouds, the baseline NSFP took 18.39/8.38 s to converge.
When replacing the 8-layer MLPs with a linear network and applying complex PE, we observed a speedup of only 1.80/2.36$\times$ in total, primarily due to the network propagation speedup. 
When we replaced naive CD loss with the DT loss, a significant 31.7/16.4$\times$ speedup was achieved while maintaining comparable high accuracy.
However, replacing a deep network with a linear model when using DT loss only resulted in an additional 1.2/1.2$\times$ speedup at the cost of a decrease in flow accuracy.
Specifically, the total computation time of our proposed method with DT loss is 0.58/0.51 s, and 
with DT loss and a linear model is 0.49/0.43 s.
However, with a marginal decrease in computation time, we observed a relatively large drop in performance such that the strict accuracy of our method with DT is 84.73/80.05\% and 71.27/65.00\% with DT and a linear model.
These results strongly support our argument in the previous section, which is the general strategy of simplifying or replacing network architectures to accelerate coordinate networks is not particularly effective when optimizing scene flow through coordinate networks.

Overall, the performance of our method using DT loss has on-par performance compared to NSFP while being orders of magnitude faster.
We have noticed that there is an ``improved accuracy'' of our method compared to NSFP in~\cref{tab:mean_time_waymo_3d},~\cref{tab:mean_time_argoverse_3d} based on these facts:
1. The performance of our method and NSFP is similar---our method achieves slightly higher performance on some metrics (\eg, accuracy), and NSFP achieves slightly higher performance on other metrics (\eg, angular error).
These results indicate that distance transform (DT) is an effective and efficient surrogate for the Chamfer distance (CD) in the scene flow problem.
2. Compared to the ``exact'' point distance (CD), DT queries the distance based on a voxelized DT map, which naturally smooths the flow estimation. 
In real-world applications, we believe using a DT-accelerated deep network model will achieve both high-fidelity accuracy and efficient computation.

\vspace{-0.35cm}
\paragraph{Computation time breakdown.}
A detailed time breakdown is provided in both tables, including pre-computation, correspondence search$\hphantom{.}$/$\hphantom{.}$DT distance query, and network forward$\hphantom{.}$/$\hphantom{.}$backward propagation.
The naive Chamfer distance requires per-point correspondence search, which can be extremely slow (43.1/17 ms per iteration), especially when the point cloud is denser.
For instance, correspondence search is much slower in the Waymo Open dataset than in the Argoverse dataset since Waymo Open has an average of 90k points while Argoverse has an average of 50k points in a single point cloud.
DT query is a pre-defined table lookup that is exceptionally faster than the naive point correspondence search, achieving a remarkable 172/71$\times$ speedup per optimization step---the main contributor to the overall efficiency of our method.
Note that the pre-computing of the DT map ($\sim$40 ms) is acceptable given that it is one-time computing and the total computation time is in the range of hundreds of milliseconds.
Additional complex PE and a linear model only provide a modest speedup of 1.6/1.5$\times$ in the network propagation step.
Further optimization of the pre-computation and faster implementation can be achieved with full CUDA support.
Note that although in NSFP, the CD loss was implemented using PyTorch3D with CUDA acceleration, it still requires significant computation time, indicating its inherent limitations that cannot be easily overcome through engineering techniques alone.

\begin{figure*}[t]
    \centering
    \includegraphics[width=\linewidth]{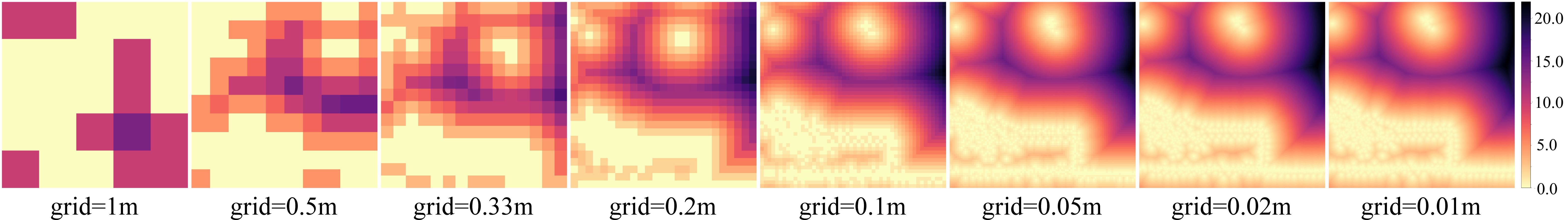}
    \caption{2D DT map with different grid sizes. A larger grid size results in extremely coarse DT, as the grid cell size decreases, the DT accuracy increases. 
    Here we zoom in on the original DT map in~\cref{fig:dt_map} to show the detailed distance values.}
    \label{fig:dt_grid_size}
    \vspace{-0.3cm}
\end{figure*}

\vspace{-0.35cm}
\paragraph{OOD generalizability.}
We further extend our model using fewer points (8,192 points) to accommodate for a fair comparison against learning-based methods~\cite{wu2020pointpwc, kittenplon2020flowstep3d, wei2021pv, puy20flot, gojcic2021weakly}.
For these learning models, a fixed number of points is required and they only operate on fewer points, such as 2,048~\cite{liu2019flownet3d, puy20flot} or 8,192 points~\cite{puy20flot, wu2020pointpwc, gojcic2021weakly, kittenplon2020flowstep3d}.
Also, these learning methods need to crop the point cloud to a small range. 
Therefore, learning models cannot be easily adapted to large-scale dense point clouds. 
Jund~\etal~explored to inference dense point cloud~\cite{jund2021scalable}, but they did not directly process the full point cloud as the input.
A simple solution to address this challenge is to iteratively predict scene flow for small subsets of dense point clouds using learning methods. 
However, this approach requires intensive computation, and it might cause out-of-memory issues~\cite{jund2021scalable}.

Fully supervised learning methods FlowStep3D, FLOT, PV-RAFT, and self-supervised method PointPWC-Net do not generalize to new autonomous driving datasets (see~\cref{tab:mean_time_waymo_3d}, ~\cref{tab:mean_time_argoverse_3d}) due to the domain gap between the training and testing datasets~\cite{pontes2020scene, li2021neural, najibi2022motion, dong2022exploiting, jin2022deformation}.
We provide additional performance in the supplementary material.
Despite its poor performance on OOD data, FLOT achieves a competitive inference time, making it one of the fastest learning-based methods available.
Note that since the weakly supervised learning method R3DSF takes supervision from object segmentation and AV ego motions and was trained on a lidar point cloud-based dataset, it has a relatively high accuracy compared to full/self-supervised learning methods, but its performance is still inferior to non-learning-based methods NSFP and our method.
Such evidence strongly suggests that our method---a runtime optimization---is robust to OOD effects, and can be directly applied to applications where no training data is readily available.

\begin{figure}[t]
    \centering
    \includegraphics[width=0.91\linewidth]{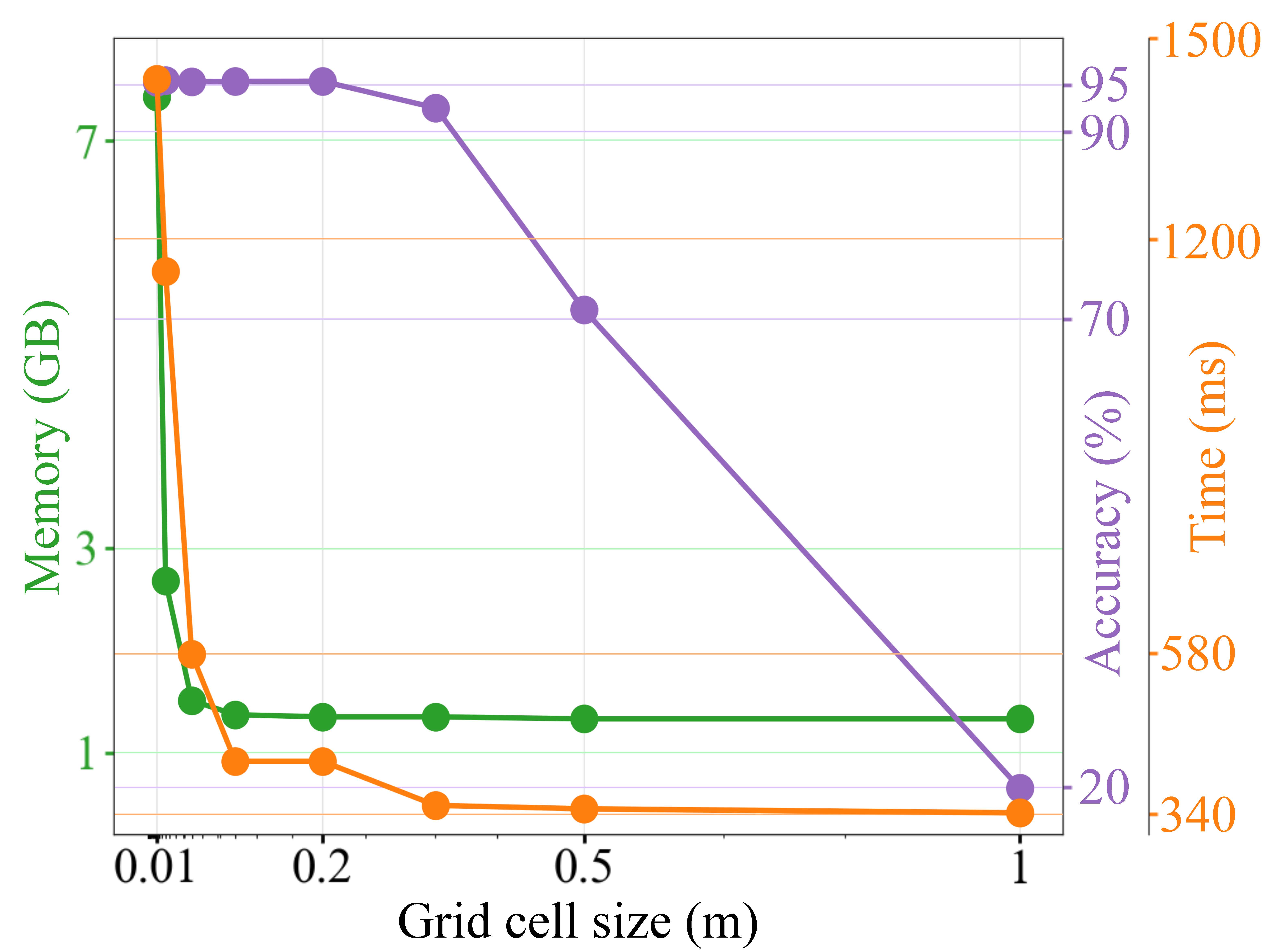}
    \vspace{0.1cm}
    \caption{Performance of distance transform with different grid cell sizes on a 2D scene from Argoverse scene flow dataset. 
    Generally, the larger the grid cell is, the lower the accuracy becomes. 
    The computation time and memory consumption also decrease as the grid cell size increases (the number of grids decreases).
    The total grid of the DT map is of size 160$\times$150.}
    \label{fig:grid_memory_time_acc}
    \vspace{-0.3cm}
\end{figure}

\vspace{-0.35cm}
\paragraph{Towards real-time computation.}

Our method attains real-time performance (121/124 ms) which is comparable to learning methods and maintains high accuracy, while learning-based methods struggle to generalize on OOD data.
It indicates the significant potential of applying robust and accurate runtime optimization-based methods in many vision-based applications.

\vspace{-0.35cm}
\paragraph{Other learning methods.}
Learning-based scene flow methods are usually fast and achieve good accuracy when applied to in-domain small-scale data (training and testing are on the same dataset with a specific range:~\eg, KITTI data with point clouds within 35m of the scene center) with a limited number of points (usually 2,048$\hphantom{.}$/$\hphantom{.}$8,192 points).
As discussed in~\cite{pontes2020scene, li2021neural, najibi2022motion, dong2022exploiting, jin2022deformation}, the domain gap is a significant challenge for learning methods---a specific dataset with specific configurations---\eg, coordinate systems, viewing directions,~\etc---that match the testing data needs to be used during training.
However, we are interested in exploring runtime optimization-based methods that are robust to large-scale OOD data that can be employed in many real-world applications, such as autonomous driving scenarios, where no labels are readily available.

\subsection{Ablation study of DT grid size}\label{sc:exp:ablation}
DT splits the space into small grid cells, while the grid size affects the accuracy, the computation time, and the memory accuracy.
We provide a performance comparison using different grid cell sizes of DT in~\cref{fig:grid_memory_time_acc}.
A 2D visualization of DT with different grid sizes is also shown in~\cref{fig:dt_grid_size}.
We can clearly see that a relatively small grid cell size ($grid\leq0.1m$) is required to ensure the fidelity of the distance transform map.
Note that when rasterized points are closer to the original irregular points, DT is a closer approximation to the exact Euclidean distance between two point clouds.
Moreover, with a rasterized point-based DT strategy, the pre-computation of the map is no longer an overhead, and the grid cell size will largely affect the memory instead of the computation time---we trade memory consumption with computation time.

\begin{table}[t]
    \caption{Comparison of our method and NSFP++ on Argoverse scene flow dataset with ego-motion compensation.}
    \centering
    \begin{adjustbox}{width=0.9\linewidth}
    \begin{tabular}{@{}lcccccc@{}}
    \toprule
    \thead{Method} & \thead{$\mathcal{E}\downarrow$ \\ $(m)$} &\thead{$Acc_5\uparrow$ \\ $(\%)$} &\thead{$Acc_{10}\uparrow$ \\ $(\%)$} & \thead{$\theta_{\epsilon}\downarrow$ \\ $(rad)$} &\thead{t $\downarrow$ \\ $(ms)$}
    \\ \midrule
    Ours & 0.411 & \textbf{34.94} & 46.82 & 0.731 & \textbf{335} \\
    NSFP++~\cite{najibi2022motion} & \underline{0.295} & \underline{31.82} & \textbf{62.61} & \textbf{0.343} & 16188 \\
    NSFP++ (DT) & \textbf{0.272} & 30.26 & \underline{60.25} & \underline{0.365} & \underline{2001} \\
    \bottomrule
    \end{tabular}
    \end{adjustbox}
    \label{tb:compare_nsfp++}
    \vspace{-0.2cm}
\end{table}

\subsection{Speedup in other methods}
\label{sc:exp:compare_nsfp++}
The proposed DT loss is a \emph{general} loss that can be used in other 3D geometry tasks.
Specifically, for optimization-based methods such as RSF~\cite{deng2023rsf}, DT could replace the nearest neighbor distance loss to speed up the overall computation.
For learning methods such as PointPWCNet~\cite{wu2020pointpwc}, DT could replace Chamfer loss to speed up optimization during training, especially when dealing with a large number of points.
For methods using cycle consistency, when replacing CD ($\bigO({N^2})$) with DT, the forward flow computation will be substantially speeded up (similar speedup to the paper).
The backward flow computation will also be speeded up even though the DT map will be computed in every optimization step.

We provide an example of comparing the results of our method to NSFP++~\cite{najibi2022motion} in~\cref{tb:compare_nsfp++}.
We used our version of NSFP++ due to no publicly available official implementation.
Since NSFP++ requires ego-motion compensation, we also created an additional Argoverse scene flow dataset that removes the ego motions of autonomous vehicles.
These ego motions are provided in the original Argoverse dataset~\cite{chang2019argoverse}.

Following~\cite{najibi2022motion}, we define dynamic points as points that the norm between the deformed source (\ie, source point cloud deformed by the ground truth flow) and the ego-motion compensated source (\ie, the source point cloud is transformed by the ego-motion) exceeds a threshold of $0.05$.
All the metrics are computed only on dynamic points.
During optimization, for our method, we removed the ego motion of the point cloud and estimated the scene flow using the full point cloud, and for NSFP++, we only used dynamic points as required.

In~\cref{tb:compare_nsfp++}, we show that our method achieves worse performance than NSFP++ while being $\sim$50$\times$ faster.
We further replaced the Chamfer distance loss used in NSFP++ with our proposed distance transform (DT) loss, and we observed a $\sim$8$\times$ speedup.
The distance transform loss can be a robust and efficient surrogate to Chamfer distance loss in many deep geometry vision tasks.

\begin{figure}[t]
    \centering
    \includegraphics[width=\linewidth]{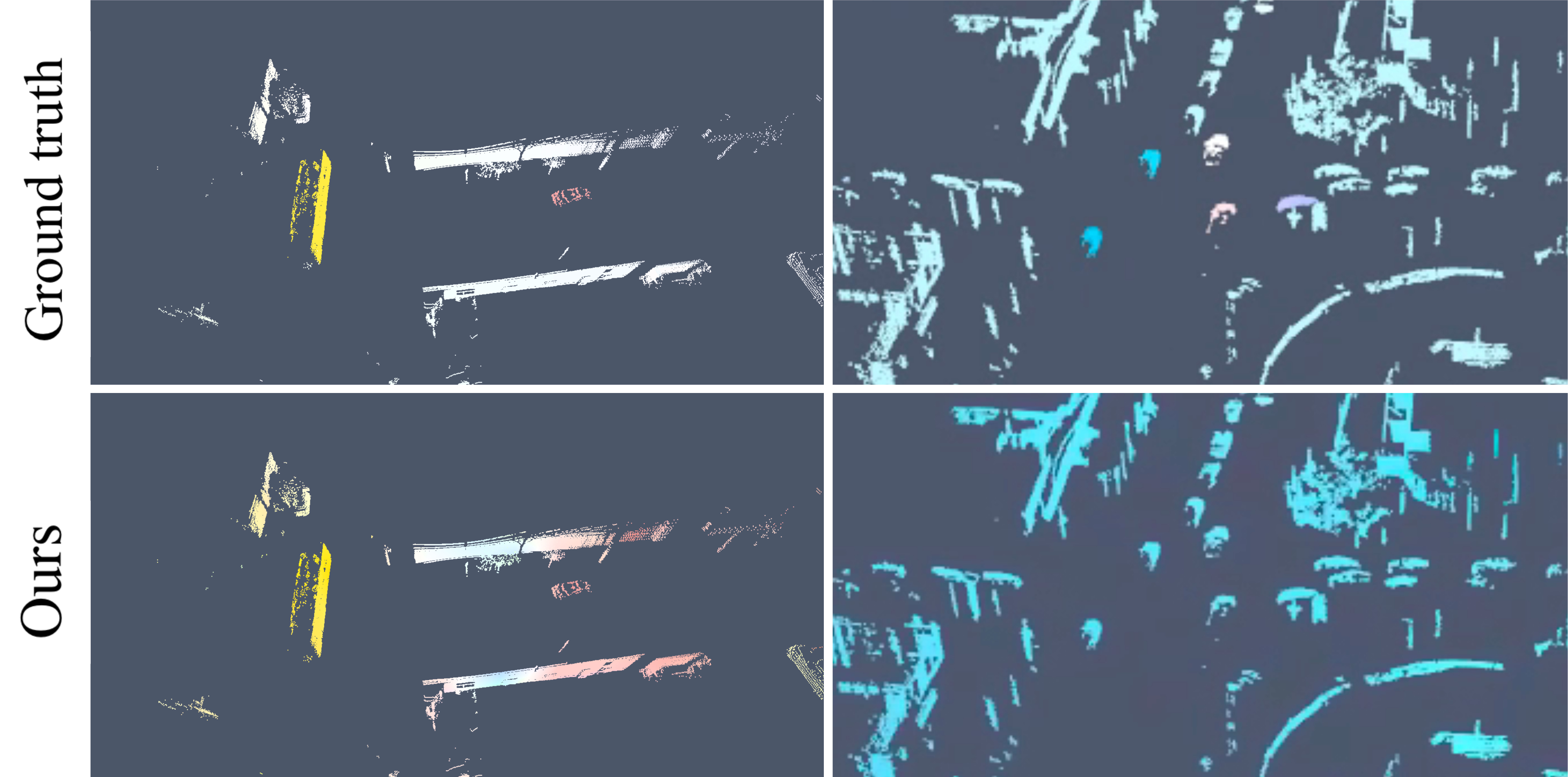}
    \caption{Failure cases on Argoverse dataset. 
    The left figure shows some noisy motions were predicted in the background. 
    The right figure shows our method failed to predict when the motion of some cars is relatively small to the AV.}
    \label{fig:argo_failure}
    \vspace{-0.35cm}
\end{figure}

\subsection{Limitations}\label{sc:exp:limitation}

One drawback of our method is that creating a DT map using rasterization can lead to discretization errors, especially when the grid size of the DT map is large.
To mitigate these, it is necessary to build a DT map with relatively fine-resolution grids.
In this case, the memory consumption will increase especially when dealing with high-dimensional data, such as 3D point clouds.
Further engineering efforts of pre-creating an efficient high-resolution DT map are required to maintain a reasonable memory cost.
However, we empirically find that in the context of scene flow estimation---specifically for scene flow in autonomous driving scenarios---a relatively smooth representation is preferred for local rigidity assumptions.
Nevertheless, the tradeoff between the grid resolution, the memory consumption, and the estimation accuracy should be carefully considered and chosen based on specific real-world applications.

\vspace{-0.3cm}
\paragraph{Failure cases.}
We show two typical failure cases in~\cref{fig:argo_failure}.
The first case (left column) is when dynamic points are much sparser than the background, our prediction can result in noisy non-rigid motions in the background. 
The second case (right column) is when the dynamic motion is relatively small, our model 
may fail to recognize the dynamic scene and only predict rigid motions.

\section{Conclusion}
In this work, we revisit the runtime optimization-based scene flow method NSFP and propose a method that is both efficient and generalizable to large-scale OOD data and dense lidar points.
We identify that common strategies for speeding up network architectures do not yield significant time reductions---the major computation overhead is the Chamfer distance loss.
Therefore, we propose to use an efficient correspondence-free distance transform loss as a robust surrogate. 
The rediscovery of DT in scene flow estimation opens up an innovative venue to leverage its efficiency and robustness for various deep geometry tasks.
Compared to NSFP, our method maintains comparable accuracy but gains up to $\sim$30$\times$ speedups. 
We report for the first time a real-time performance ($\sim$120 ms) with neural scene flow and runtime optimization when using fewer points (8,192).
The efficient runtime optimization-based neural scene flow can be widely applied in lidar scenes to do point cloud densification, open-world object detection, and scene clustering, such as in autonomous driving scenarios, where no ground truth or training data are readily available.

\vspace{0.3cm}
\noindent\textbf{Acknowledgement:}\:
We would like to thank Haosen Xing for the careful review of the manuscript and the help throughout the project. We thank Kavisha Vidanapathirana for the initial implementation of NSFP++~\cite{najibi2022motion}.

{\small
\bibliographystyle{ieee_fullname}
\bibliography{egbib}
}

\newpage

\appendix
\section*{\Large Appendix}

\section{Experiments}
\label{supp:exp}

\subsection{Speedup in network architectures}
\label{supp:speedup_architecture}

\paragraph{Positional encodings.}
Positional encodings (PEs) are usually used in coordinate networks~\cite{mildenhall2020nerf} to increase the bandwidth of the input by multiple (random) frequencies. 
With PEs, the coordinate network converges much faster and achieves better performance~\cite{tancik2020fourier}. 
Moreover, a recent paper~\cite{zheng2022trading} points out that the success of PEs is attributed to the rank increase of the input: the deepness of the neural network is to increase the rank of the embedding and aligns the embedding space to the output space. 
However, if the rank of the input is higher, the network can be shallower. 
Therefore,~\cite{zheng2022trading} uses a more complex PE to dramatically increase the rank of the input, thus the followed deep non-linear network can be replaced by a shallow linear function.

\paragraph{Complex PE-based linear model.}
\label{supp:complex_encoding}
In our paper, we implemented a complex positional encoding (PE)~\cite{zheng2022trading}-based linear model to test the efficiency of simplifying network architectures.
While simple PE refers to a simple concatenation of the encoding in each input dimension, complex PE~\cite{zheng2022trading} is a more complicated encoding that computes the Kronecker product of the per-axis encoding.
As mentioned in~\cite{zheng2022trading}, one reason behind the success of the deep network is that it increases the rank of the low-rank input.
Therefore, if the input to the network has a high rank, the network can be shallower accordingly.
To increase the rank of the input, we use a complex encoding instead of a deep network.
The rank of the complex encoding is
\begin{align}
    \mbox{Rank} & \left(\phi(\mathbf{p}_x) \otimes \phi(\mathbf{p}_y) \otimes \phi(\mathbf{p}_z) \right) = \notag \\
    & \mbox{Rank} \left(\phi(\mathbf{p}_x) \right) \mbox{Rank} \left(\phi(\mathbf{p}_y) \right) \mbox{Rank} \left(\phi(\mathbf{p}_z) \right), \label{supp:eq:rank_theory}
\end{align}
which achieves full rank that allows us to only use a linear layer $\mathbf{W}$ as a follow-up embedder.

The advantage of a complex PE lies in two aspects: first, with a linear layer, the problem can be solved analytically in many cases; 
second, if the closed-form solution is difficult to obtain, using a linear layer in iterative solvers, such as gradient descent-based methods, will converge faster. 
Similar to frequency-based encodings~\cite{tancik2020fourier}, shift-based encodings like Gaussian or Triangle wave also work similarly well~\cite{zheng2022trading}. 
These shift-based encodings involve very few parameters for each sample point due to sparsity, while a deep network requires significant amounts of parameters.

To reconstruct the signal $\mathbf{S}$, we optimize
\begin{align}
    \argmin_{\mathbf{W}} \left\| \mbox{vec}(\mathbf{S}) {-} \left(\phi(\mathbf{p}_x) \otimes \phi(\mathbf{p}_y) \otimes \phi(\mathbf{p}_z) \right)^T {\mbox{vec}(\mathbf{W})} \right\|_2^2. \label{supp:eq:closed-form}
\end{align}
When the coordinate is separable along each axis (\eg, 2D image), using Kronecker product, we have a closed-form solution as
\begin{align}
    \mathbf{W} = \phi(\mathbf{p}_x)^{-1} \mathbf{S} \phi(\mathbf{p}_y)^{-T} \phi(\mathbf{p}_z)^{-T}. \label{supp:eq:closed-form-kronecker}
\end{align}

With all the complex PE theory in hand, it is nontrivial to implement it in a scene flow problem, and to the best of our knowledge, we are the first to apply complex PE to real-world large-scale data. 
To employ complex PE in the scene flow problem, we first replace the non-linear multi-layer perceptrons (MLPs) in NSFP with a linear layer parameterized by $\mathbf{W} \:{\in}\: \mathbb{R}^{W_xW_yW_z \times 3}$---$W_x$, $W_y$, and $W_z$ are encodings in each dimension and $3$ is the dimension of the flow---and encode the input coordinates in complex PE form. 
The flow represented by MLPs can then be modified as
\begin{align}
    \mathbf{f} \;{=}\; g \left(\mathbf{p};\: \mathbf{W} \right) = \left(\phi(\mathbf{p}_z) \otimes \phi(\mathbf{p}_y) \otimes \phi(\mathbf{p}_x) \right) \mbox{vec}\left(\mathbf{W}\right) \label{supp:eq:complex_encoding}
\end{align}
where $\phi\left(\cdot\right)$ is the encoder, $\mathbf{p}_x$, $\mathbf{p}_y$, $\mathbf{p}_z$ are the sample points in $x$, $y$, $z$ coordinates.

\paragraph{Blending function.}
However, in the 3D point cloud case, the unordered points are not separable on each axis.
A blending function $\mbox{B}$ is introduced to interpolate the points and avoid the computation of large naive complex PE. 
Please note that the unordered point cloud has non-separable coordinates.
According to~\cite{zheng2022trading}, the non-separable-coordinate problem can be approximated by a blending function and an encoding of virtual separable grid points. 
The blending approximation is $\phi(\mathbf{p}_z) \:{\otimes}\: \phi(\mathbf{p}_y) \:{\otimes}\: \phi(\mathbf{p}_x) \:{\approx}\: \mbox{B}(\mathbf{p};\phi)\phi(\mathbf{z}) \:{\otimes}\: \phi(\mathbf{y}) \:{\otimes}\: \phi(\mathbf{x})$, where $\mathbf{x} \:{\in}\: \mathbb{R}^{W_x}$, $\mathbf{y} \:{\in}\: \mathbb{R}^{W_y}$ and $\mathbf{z} \:{\in}\: \mathbb{R}^{W_z}$ are virtual grid points.
Using such approximation, the computation of complex PE of grid points is as easy and fast as a matrix multiplication due to the property of the Kronecker product. 
Intuitively, the blending matrix $\mbox{B}$ can be viewed as a matrix consisting of non-linear interpolation coefficients that depend on the encoding function $\phi(\cdot)$. 
It is large but sparse,~\ie, there are only 8 non-zero values on each row (corresponding to the 8 neighboring grid points of the query point), we can index the matrix efficiently by only querying the non-zero entries. 
Meanwhile, different from~\cite{zheng2022trading}, grids here have physical meanings and their size can be adjusted.

Therefore, the scene flow becomes
\begin{align}
    \mathbf{f} \approx \mbox{B}(\mathbf{p};\phi) \mbox{vec} \left(\phi(\mathbf{x}) \mathbf{W} \phi(\mathbf{z})^T \phi(\mathbf{y})^T \right), \label{supp:eq:approx_blending}
\end{align}
where $\phi(\mathbf{x}) \mathbf{W} \phi(\mathbf{z})^T \phi(\mathbf{y})^T$ is a simple notation for $n$-mode multiplication.
And the scene flow optimization is
\begin{align}
    {\argmin_{\mathbf{W}}} \left\| \mathbf{f} {-} {\mbox{B}} \left({\phi(\mathbf{p}_x)} \otimes {\phi(\mathbf{p}_y)} \otimes {\phi(\mathbf{p}_z)} \right)^T {\mbox{vec}(\mathbf{W})} \right\|_2^2. \label{supp:eq:b_optimization}
\end{align}
We therefore solve $\mathbf{W}$ using gradient descent and distance transform loss as
\begin{align}
    \mathbf{W}^* {=} & \argmin_{\mathbf{W}} {\sum_{\mathbf{p} \in \mathcal{S}_1}} \mbox{D} \left( \mathbf{p} {+} \mbox{B} \left( \mathbf{p}; \phi \right) \right. \notag \\ 
    & \left. \mbox{vec} \left(\phi(\mathbf{x}) \mathbf{W} \phi(\mathbf{z})^T \phi(\mathbf{y})^T \right), \mathcal{S}_2 \right). \label{supp:eq:optim_complex}
\end{align}

Compared to a $L$ layer network of width $W$ used in NSFP to process $N$ sample points, a linear layer of size $W_x{\times} W_y{\times} W_z$ from the Kronecker product speeds up the network significantly from $\mathcal{O}{(NW^{2}L)}$ to  $\mathcal{O}{\left(8N{+}3W_{x}W_{y}W_{z}\left(W_{x}{+}W_{y}{+}W_{z}\right)\right)}$. 
We further constrain the optimization by applying an explicit total variation (TV) regularizer on $\mathbf{W}$ as:
\begin{align}
    & \mbox{TV} \left(\mathbf{W}\right) = \frac{1}{(W_{x}{-}1)(W_{y}{-}1)(W_{z}{-}1)} \notag \\
    & {\sum_{i{=}0}^{W_{x}{-}2}} {\sum_{j{=}0}^{W_{y}{-}2}} {\sum_{k{=}0}^{W_{z}{-}2}} \sqrt{(d\mathbf{x}_{i,j,k})^{2}{+}(d\mathbf{y}_{i,j,k})^{2}{+}(d\mathbf{z}_{i,j,k})^{2}},
    \label{supp:eq:tv_loss}
\end{align}
where $\mbox{dx}_{i,j,k} \:{=}\: \mathbf{W}_{i,j,k} \:{-}\: \mathbf{W}_{i{+}1,j,k}$, $\mbox{dy}_{i,j,k} \:{=}\: \mathbf{W}_{i,j,k} \:{-}\:$ $\mathbf{W}_{i,j{+}1,k} $, $\mbox{dz}_{i,j,k} = \mathbf{W}_{i,j,k} {-}\mathbf{W}_{i,j,k{+}1} $.
In all, the loss function of the complex PE model is (with TV) 
\begin{align}
    \mathcal{L} \left(\mathbf{W} \right) = & \sum_{\mathbf{p} \in \mathcal{S}_1} \mbox{D} \left( \mathbf{p} {+} \mbox{B} \left( \mathbf{p} \right) \mbox{vec} \left(\phi(\mathbf{x}) \mathbf{W} \phi(\mathbf{z})^T \right.\right. \notag \\
    & \left.\left. \hphantom{DD} \phi(\mathbf{y})^T \right), \mathcal{S}_2 \right) + \frac{\lambda}{2} \mbox{TV} \left(\mathbf{W}\right). \label{supp:eq:loss_f_g}
\end{align}

\subsection{Speedup in point correspondence search} 
\label{supp:speedup_chamfer}

Before the deployment of the distance transform (DT), we explored other strategies to speed up the point correspondence search in Chamfer distance.

\paragraph{Build a k-d tree to search nearest neighboring points.}
The point distance function $\mbox{D}$ is computationally intensive, as a set of point-to-point correspondences needs to be optimized in each optimization step.
One speedup is to construct a k-d tree to accelerate the nearest neighbor search which reduces the computation complexity of point correspondence search from $\mathcal{O}{({n^2})}$ to $\mathcal{O}{(n{\log n})}$.
Since the target point cloud $\mathcal{S}_2$ is fixed, we only need to pre-build the k-d tree for $\mathcal{S}_2$ once. 
However, the source point cloud $\mathcal{S}_1$ is deformed in each optimization step, making the pre-build of the source point cloud k-d tree happen in every iteration, not to mention that the per-iteration k-d tree query is another computation overhead when the number of points is big.

\begin{table*}[t]
\caption[]{\textbf{Additional computation time and performance on Waymo Open Scene Flow dataset.} 
The upper tabular between {\color{sol_blue}\textbf{blue bars}} are experiments with the full point cloud, and the lower tabular between {\color{beer_orange}\textbf{orange bars}} are experiments with only 8,192 points.
Corr. / k-d tree / DT query denotes correspondence search, k-d tree-based correspondence search, or DT query.
}
    \centering
    \begin{adjustbox}{width=\linewidth}
    \begin{tabular}{@{}clcccccrccccc@{}}
        \toprule
        &\multirow{2}{*}{\thead{\normalsize Method}}
        &\multirow{2}{*}{\thead{\normalsize ${\mathcal{E}}$\\${(m)}\downarrow$}} 
        &\multirow{2}{*}{\thead{\normalsize ${Acc_5}$\\${(\%)}\uparrow$}} 
        &\multirow{2}{*}{\thead{\normalsize ${Acc_{10}}$\\${(\%)}\uparrow$}} 
        &\multirow{2}{*}{\thead{\normalsize ${\theta_{\epsilon}}$\\${(rad)}\downarrow$}} 
        &\multicolumn{4}{c}{\thead{\normalsize $t~  (ms)\downarrow$}} \\
        \cmidrule{7-10}
        &&&&&& \thead{\normalsize Pre-compute} &\thead{\normalsize Corr. / k-d tree / DT query} & Network &\thead{\normalsize Total} \\
        \midrule
        \arrayrulecolor{sol_blue}\toprule[0.5ex]
        & NSFP (baseline) & 0.118 & 74.16 & 86.70 & 0.300 & --- 
        & \boxitred{3.5in} $\hphantom{.}$ 43.1 [15036] $\hphantom{\vert \vert \vert {\color{tabgreen}\mathbf{20{\times}}}}$
        & 2.38 [904] $\hphantom{\vert \vert \vert .{\color{tabgreen}\mathbf{0{\times}}}}$
        & 18.39 s $\hphantom{\vert 8 8\vert {\color{tabgreen}\mathbf{20{\times}}}}$ \\
        & Baseline (k-d tree CD) & 0.104 & 74.13 & 86.81 & 0.296 
        & 15.24 [5283] $\vert \hphantom{\vert} {\color{tabgreen}\mathbf{2.8{\times}}}$
        & $\hphantom{...}$ 2.27 [838] $\vert {\color{tabgreen}\mathbf{1.05{\times}}}$ $\hphantom{.}$ 
        & 8.51 s $\vert {\color{tabgreen}\mathbf{2.16{\times}}}$ \\
        & Baseline (k-d tree CD, linear) & 0.101 & 70.14 & 86.24 & 0.315 
        & 12.92 [2349] $\vert \hphantom{\vert} {\color{tabgreen}\mathbf{3.3{\times}}}$
        & $\hphantom{||}$ 1.39[262] $\vert {\color{tabgreen}\mathbf{1.71{\times}}}$ 
        & 4.15 s $\vert {\color{tabgreen}\mathbf{4.43{\times}}}$ \\
        \arrayrulecolor{sol_blue}\toprule[0.5ex]
        \arrayrulecolor{beer_orange}\toprule[0.5ex]
        & PointPWC-Net~\cite{wu2020pointpwc} & 4.109 & 0.05 & 0.36 & 1.742 & --- 
        & --- $\hphantom{----}$ & --- 
        & 185 ms $\vert {\color{tabred}\mathbf{1.32{\times}}}$ $\hphantom{.}$ \\
        & FlowStep3D~\cite{kittenplon2020flowstep3d} & 0.753 & 0.01 & 0.09 & 1.212 & --- & --- $\hphantom{----}$ & --- 
        & 725 ms $\vert {\color{tabred}\mathbf{5.18{\times}}}$ $\hphantom{.}$ \\
        & PV-RAFT~\cite{wei2021pv} & 10.675 & 0.03 & 0.13 & 1.794 & --- & --- $\hphantom{----}$ & --- 
        & 505 ms $\vert {\color{tabred}\mathbf{3.61{\times}}}$ $\hphantom{.}$ \\
        & R3DSF~\cite{gojcic2021weakly} & 0.414 & 35.47 & 44.96 & 0.527 & --- & --- $\hphantom{----}$ & --- 
        & \boxitgreen{1.01in} \underline{140 ms} $\hphantom{\vert {\color{tabred}\mathbf{1.11{\times}}}}$ $\hphantom{.}$ \\
        & Ours (8,192 pts) & \textbf{0.106} & \textbf{77.53} & \textbf{88.99} & \textbf{0.329} & 35.22 
        & 0.23 [6.5] $\vert {\color{tabgreen}\mathbf{496{\times}}}$
        & $\hphantom{||}$ 2.60 [76] $\vert {\color{tabgreen}\mathbf{1.8{\times}}}$
        & \textbf{121 ms} $\vert {\color{tabgreen}\mathbf{1.16{\times}}}$ $\hphantom{.}$ \\
        \arrayrulecolor{beer_orange}\toprule[0.5ex]
        \arrayrulecolor{black}\bottomrule
    \end{tabular}
    \end{adjustbox}
    \label{supp:tab:mean_time_waymo_3d}
\end{table*}

\begin{table*}[t]
\caption[]{\textbf{Additional computation time and performance on Argoverse Scene Flow dataset.}
}
    \centering
    \begin{adjustbox}{width=\linewidth}
    \begin{tabular}{@{}clcccccrccccc@{}}
        \toprule
        &\multirow{2}{*}{\thead{\normalsize Method}}
        &\multirow{2}{*}{\thead{\normalsize ${\mathcal{E}}$\\${(m)}\downarrow$}} 
        &\multirow{2}{*}{\thead{\normalsize ${Acc_5}$\\${(\%)}\uparrow$}} 
        &\multirow{2}{*}{\thead{\normalsize ${Acc_{10}}$\\${(\%)}\uparrow$}} 
        &\multirow{2}{*}{\thead{\normalsize ${\theta_{\epsilon}}$\\${(rad)}\downarrow$}} 
        &\multicolumn{4}{c}{\thead{\normalsize $t~(ms)\downarrow$}} \\
        \cmidrule{7-10}
        &&&&&& \thead{\normalsize Pre-compute} &\thead{\normalsize Corr. / k-d tree / DT query} & Network &\thead{\normalsize Total} \\
        \midrule
        \arrayrulecolor{sol_blue}\toprule[0.5ex]
        & NSFP (baseline) & 0.078 & 69.46 & 86.22 & \underline{0.253}
        & --- 
        & \boxitred{3.5in} $\hphantom{....}$ 17 [5901] $\hphantom{\vert \vert \vert {\color{tabgreen}\mathbf{20{\times}}}}$
        & 2.31 [848] $\hphantom{\vert \vert {\color{tabgreen}\mathbf{20{\times}}}}$
        & 8.38 s $\hphantom{\vert 8 \vert {\color{tabgreen}\mathbf{20{\times}}}}$ \\
        & Baseline (k-d tree CD) & 0.078 & 69.14 & 85.99 & 0.253 
        & 11.3 [4063] $\vert \hphantom{\vert} {\color{tabgreen}\mathbf{1.5{\times}}}$ 
        & 2.28 [830] $\vert {\color{tabgreen}\mathbf{1.0{\times}}}$ 
        & 6.25 s $\vert {\color{tabgreen}\mathbf{1.34{\times}}}$ \\
        & Baseline (k-d tree CD, linear) & 0.071 & 68.72 & 86.39 & 0.288 
        & 9.36 [1701] $\vert \hphantom{.} {\color{tabgreen}\mathbf{1.8{\times}}}$ 
        & 1.41 [253] $\vert {\color{tabgreen}\mathbf{1.6{\times}}}$ 
        & 3.09 s $\vert {\color{tabgreen}\mathbf{2.71{\times}}}$ \\
        \arrayrulecolor{sol_blue}\toprule[0.5ex]
        \arrayrulecolor{beer_orange}\toprule[0.5ex]
        & PointPWC-Net~\cite{wu2020pointpwc} & 5.600 & 0.03 & 0.18 & 1.179 & --- & --- $\hphantom{----}$ & --- & 186 ms $\vert {\color{tabred}\mathbf{1.65{\times}}}$ $\hphantom{.}$ \\
        & FlowStep3D~\cite{kittenplon2020flowstep3d} & 0.845 & 0.01 & 0.08 & 1.860 & --- & --- $\hphantom{----}$ & --- & 729 ms $\vert {\color{tabred}\mathbf{6.45{\times}}}$ $\hphantom{.}$ \\
        & PV-RAFT~\cite{wei2021pv} & 10.745 & 0.02 & 0.10 & 1.517 & --- & --- $\hphantom{----}$ & --- & 504 ms $\vert {\color{tabred}\mathbf{4.46{\times}}}$ $\hphantom{.}$ \\
        & R3DSF~\cite{gojcic2021weakly} & 0.417 & 32.52 & 42.52 & 0.551 & --- & --- $\hphantom{----}$ & --- 
        & \boxitgreen{1.01in} \textbf{113 ms} $\hphantom{\vert {\color{tabred}\mathbf{1.41{\times}}}}$ $\hphantom{.}$ \\
        & Ours (8,192 pts) & \underline{0.118} & \textbf{69.93} & \textbf{83.55} & \underline{0.352} & 41.57 
        & 0.22 [6.33] $\vert {\color{tabgreen}\mathbf{214{\times}}}$
        & 2.51 [72.69] $\vert \hphantom{\vert} {\color{tabgreen}\mathbf{1.9{\times}}}\hphantom{.}$
        & \underline{124 ms} $\vert {\color{tabred}\mathbf{1.10{\times}}}$ $\hphantom{.}$ \\
        \arrayrulecolor{beer_orange}\toprule[0.5ex]
        \arrayrulecolor{black}\bottomrule
    \end{tabular}
    \end{adjustbox}
    \label{supp:tab:mean_time_argoverse_3d}
\end{table*}

\paragraph{Randomly sample points.}
Stochastic gradient descent (SGD)~\cite{robbins1951stochastic} is now broadly used in large-scale learning problems. 
It approximates the actual gradient descent by only computing the gradient of a randomly selected subset of the original dataset at each iteration. 
However, it can achieve relatively faster updates and guarantee a satisfied global convergence, especially when dealing with large-scale high-dimensional optimization~\cite{bottou2018optimization}.

Inspired by the idea of SGD, we choose to randomly subsample points of the dense point cloud at each iteration. 
Analogous to SGD, each individual point is viewed as sample data.
A naive sampling strategy is to sample a fixed number of points. 
Instead, we develop sampling strategies based on the number of iterations or the decreasing percentage of the loss function.
We sample fewer points at the beginning of the optimization when the point correspondences are noisy, and gradually increase the number of sampled points when the optimization becomes better constrained and finds better correspondences.

However, we also noticed that point sampling is not a practical strategy when applied to real-world problems, such as autonomous driving scenarios, where all points are needed to get sufficient information for detailed non-rigid motions.

\paragraph{Reduce the frequency of updating correspondence.}
Although~\cref{eq:optim_main} of the main paper optimizes network weights (scene flow) through an explicit point distance function, the point correspondence optimization is implicitly included. 
We have mentioned that the optimization of the point correspondence and the scene flow are highly entangled. 
We cannot easily get a good scene flow estimation even given the optimal point correspondences.

Instead of separating the scene flow and correspondence optimization, we reduce the updating frequency of the point correspondences from every single iteration to several iterations.
To guarantee a good initialization, we initially consecutively update correspondences for a fixed number of iterations.

However, the correspondence sampling strategy is unfavorable due to a considerable performance compromise.

\subsection{Implementation details}
\label{supp:implement}
We provide more implementation details for our method. Further details will be provided upon code release.

\paragraph{Datasets.}
We followed~\cite{pontes2020scene, li2021neural} to create the pseudo scene flow labels, and removed ground points according to each dataset.
Note that we used the raw point cloud from the lidar sensor and did not crop the data to a small range.

\paragraph{Truncated Chamfer distance.}
We used a truncated Chamfer distance loss for our baseline implementation as mentioned in the original NSFP~\cite{li2021neural} that is unbiased on extreme points.
Practically, we chose $2m$ as a threshold to eliminate large point distance.

\begin{figure*}[t!]
\centering
    \includegraphics[width=\linewidth]{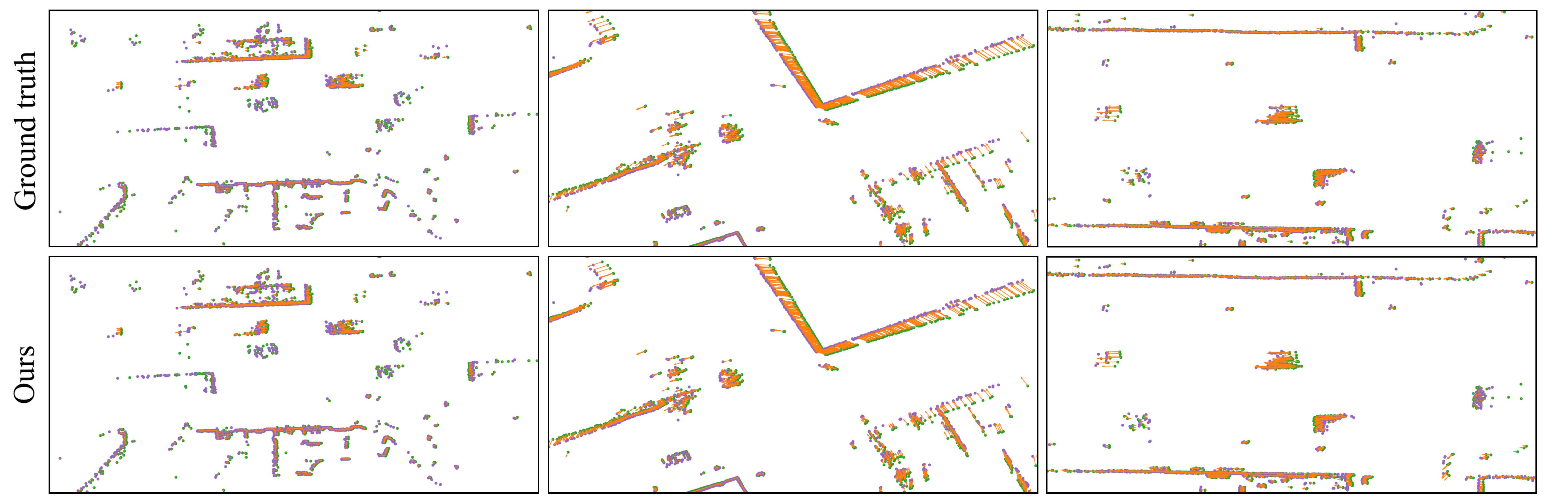}
    \caption{Visual results of the 2D scene flow estimation using our method on the Argoverse scene flow dataset. \textbf{\color{realtabgreen}Green} points are source, \textbf{\color{realtabpurple}purple} points are target, and \textbf{\color{realtaborange}orange} arrows represent the flow vectors.} 
    \label{supp:fig:bev_visual}
\end{figure*}

\paragraph{Complex positional encoding.}
Since point clouds are non-separable in 3D space, we first encoded the separable 3D virtual voxel vertices using shifted Gaussian encoders as depicted in~\cref{fig:3d_complex_encoding} of the main paper.
Since the 3D space in autonomous driving scenarios is large, we empirically found that a relatively larger voxel size (\eg, $2m$ or $5m$ for autonomous driving scene flow datasets) that constrains the motions as rigid as possible within a larger local region is more suitable to encode scene flow.
The choice of the Gaussian sigma also depends on the voxel size.
Generally, the sigma of Gaussian encoding should be twice larger than the voxel size.
For example, for a voxel size of $2m$, $\sigma>4$ is favored.
Note that the Gaussian sigma and the voxel size can be adjusted within a small range.

\subsection{Additional results}
We provide additional results on Waymo Open and Argoverse scene flow datasets in~\cref{supp:tab:mean_time_waymo_3d} and~\cref{supp:tab:mean_time_argoverse_3d} respectively.

We show how k-d tree-based correspondence search for CD loss speeds up the optimization, yet remains less effective, which indicates the inherent computation cost of correspondence search in CD loss cannot be easily solved using engineering techniques.
The performance of the linear model drops by a large margin, suggesting that it is a less favorable choice for scene flow estimation.

\subsection{Performance gap of learning methods}

The performance of learning-based methods such as PointPWC-Net~\cite{wu2020pointpwc}, FlowStep3D~\cite{kittenplon2020flowstep3d}, PV-RAFT~\cite{wei2021pv}, FLOT~\cite{puy20flot} is inferior compared to non-learning-based methods (shown in~\cref{supp:tab:mean_time_waymo_3d} and~\cref{supp:tab:mean_time_argoverse_3d}, between the {\color{beer_orange}\textbf{orange}} bar).
As discussed in the main paper, the performance gap between the learning methods and the non-learning-based methods lies in the OOD generalizability.
Even trained on similar lidar sensors---\eg, FlowStep3D was trained on the KITTI~\cite{menze2015object} dataset---the Waymo Open~\cite{sun2020scalability} and Argoverse~\cite{chang2019argoverse} scene flow datasets have different point cloud range, coordinate configurations,~\etc~to KITTI dataset, making the pre-trained model vulnerable to these data variations.
In contrast, non-learning-based methods maintained high accuracy on different datasets.
Note that our method still has competitive efficiency among these learning methods.

\begin{figure*}[t]
\centering
    \includegraphics[width=\linewidth]{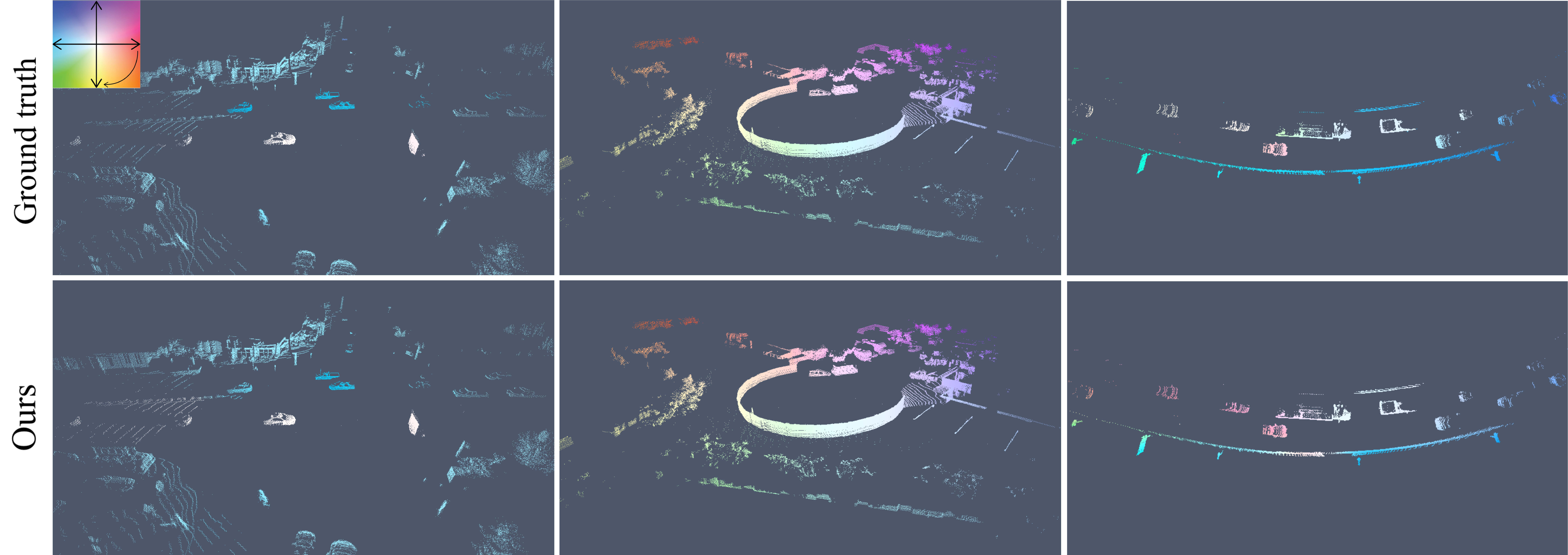}
    \caption{Visual examples of the scene flow prediction using our method on Waymo Open scene flow dataset. } 
    \label{supp:fig:waymo_visual}
\end{figure*}

\begin{figure*}[t]
\centering
    \includegraphics[width=\linewidth]{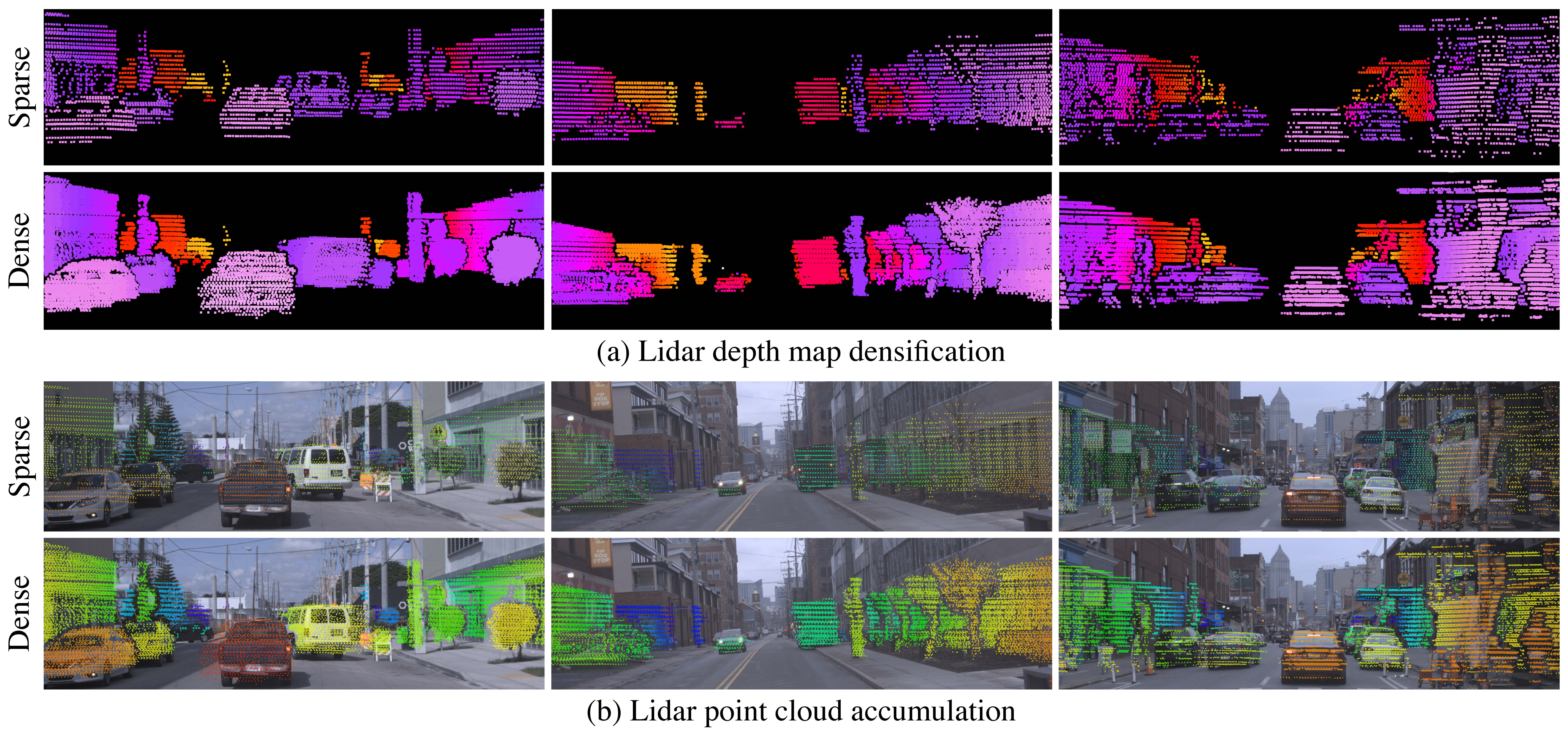}
    \caption{Visualization of the application of our method: (a) depth densification, and (b) point cloud accumulation. 
    We present the original sparse scenes on the upper row and the densified results on the lower row. 
    For point cloud accumulation, we projected the densified point cloud to the corresponding image plane for better visualization.} 
    \label{supp:fig:densification}
    \vspace{-0.25cm}
\end{figure*}

\subsection{Additional results of DT grid size}
\label{supp:exp:dt_grid_size}
\begin{table}[t]
    \caption{Performance of distance transform with different grid cell sizes on a 2D BEV scene from Argoverse scene flow dataset. 
    The total grid is of size 160$\times$150.}
    \centering
    \begin{adjustbox}{width=0.9\linewidth}
    \begin{tabular}{@{}lcccccc@{}}
    \toprule
    \thead{Grid cell \\ size (m)} & \thead{$\mathcal{E}\downarrow$ \\ $(m)$} &\thead{$Acc_5\uparrow$ \\ $(\%)$} &\thead{$Acc_{10}\uparrow$ \\ $(\%)$} & \thead{$\theta_{\epsilon}\downarrow$ \\ $(rad)$} &\thead{t $\downarrow$ \\ $(ms)$} & \thead{Mem. $\downarrow$ \\ $(GB)$}
    \\ \midrule
    1 & 0.225 & 4.30 & 19.91 & 0.189 & 342 & 1.33 \\
    0.5 & 0.123 & 34.69 & 70.98 & 0.135 & 348 & 1.33 \\
    0.33 & 0.089 & 73.25 & 92.51 & 0.116 & 353 & 1.35 \\
    0.2 & 0.071 & 90.15 & 95.36 & 0.105 & 419 & 1.35 \\
    0.1 & 0.060 & 94.31 & 95.35 & 0.097 & 419 & 1.37 \\
    0.05 & 0.059 & 94.22 & 95.31 & 0.097 & 579 & 1.51 \\
    0.02 & 0.060 & 94.26 & 95.46 & 0.095 & 1151 & 2.68 \\
    0.01 & 0.058 & 93.92 & 95.25 & 0.095 & 1438 & 7.42 \\
    \bottomrule
    \end{tabular}
    \end{adjustbox}
    \label{supp:tb:dt_grid_size}
\end{table}

We provide additional results on a 2D bird's eye view (BEV) scene in~\cref{supp:tb:dt_grid_size}.
The result is aligned with the main paper~\cref{fig:grid_memory_time_acc}.

\subsection{2D BEV visual results}
\label{supp:bev}
Some visual results of the 2D BEV scenes of the Argoverse scene flow dataset are shown in~\cref{supp:fig:bev_visual}.

\subsection{Visual results}
\label{supp:visual}

Please see~\cref{supp:fig:waymo_visual} and the project webpage \href{https://lilac-lee.github.io/FastNSF}{https://lilac-lee.github.io/FastNSF} for more visual results and applications.

\section{Application: point accumulation}
The implicit and continuous neural function allows for easy point accumulation with per-pair scene flow estimation.
Moreover, with the speedup that our method has achieved, the computation of point accumulation substantially decreased, making it possible for large amounts of point densification.

\vspace{-0.1cm}
\subsection{Continuous scene flow field}
It is important to note that using DT to replace CD will not alter the continuous property.
Therefore, similar to NSFP, our method creates a continuous flow field in that the network itself interpolates the motion of the entire 3D space, enabling long-term flow estimation and point densification through forward integration.

\vspace{-0.1cm}
\subsection{Dense point cloud accumulation}
We followed~\cite{li2021neural} to accumulate point clouds using Euler integration with per-pair scene flow estimation for the Argoverse scene flow dataset.
Different from~\cite{li2021neural}, we compute per-pair scene flow for each consecutive pair (\ie, frame 1$\rightarrow$2, frame 2$\rightarrow$3, ..., frame 10$\rightarrow$11) and interpolate fast neural scene flow to integrate 10 point clouds following the reference frame into the reference frame to densify the depth map and the point cloud.
Some visual examples are shown in~\cref{supp:fig:densification}.

\end{document}